\def\paperlanguage{}
\pdfoutput=1 

\documentclass[letterpaper, 10 pt, conference]{ieeeconf}  

\usepackage{bm}
\usepackage{cite}
\usepackage{flushend}
\include{preamble}

\IEEEoverridecommandlockouts                              

\title{\LARGE \textbf
  {
    \switchlanguage%
    {%
      Automatic Grouping of Redundant Sensors and Actuators Using Functional and Spatial Connections: Application to Muscle Grouping for Musculoskeletal Humanoids
    }%
    {%
      機能的・空間的結合を利用した冗長なセンサ・アクチュエータの自動分割: 筋骨格ヒューマノイドの筋分割への適用
    }%
  }
}

\author{Kento Kawaharazuka$^{1}$, Manabu Nishiura$^{1}$, Yuya Koga$^{1}$, Yusuke Omura$^{1}$,\\Yasunori Toshimitsu$^{1}$, Yuki Asano$^{1}$, Kei Okada$^{1}$, Koji Kawasaki$^{2}$, and Masayuki Inaba$^{1}$
  \thanks{$^{1}$ The authors are with the Department of Mechano-Informatics, Graduate School of Information Science and Technology, The University of Tokyo, 7-3-1 Hongo, Bunkyo-ku, Tokyo, 113-8656, Japan.
    {\texttt\small [kawaharazuka, nishiura, koga, oomura, toshimitsu, asano, k-okada, inaba]@jsk.t.u-tokyo.ac.jp}
  }
  \thanks{$^{2}$ The author is associated with TOYOTA MOTOR CORPORATION.
    {\texttt\small koji\_kawasaki@mail.toyota.co.jp}
  }
}

\begin{document}

\maketitle
\thispagestyle{empty}
\pagestyle{empty}

\begin{abstract}
  \switchlanguage%
  {%
    For a robot with redundant sensors and actuators distributed throughout its body, it is difficult to construct a controller or a neural network using all of them due to computational cost and complexity.
    Therefore, it is effective to extract functionally related sensors and actuators, group them, and construct a controller or a network for each of these groups.
    In this study, the functional and spatial connections among sensors and actuators are embedded into a graph structure and a method for automatic grouping is developed.
    Taking a musculoskeletal humanoid with a large number of redundant muscles as an example, this method automatically divides all the muscles into regions such as the forearm, upper arm, scapula, neck, etc., which has been done by humans based on a geometric model.
    The functional relationship among the muscles and the spatial relationship of the neural connections are calculated without a geometric model.
    This study is applied to muscle grouping of musculoskeletal humanoids Musashi and Kengoro, and its effectiveness is verified.
  }%
  {%
    冗長なセンサ・アクチュエータが全身に分布するロボットにおいて, それら全てを用いて一つの制御器やニューラルネットワークを構築することは, 計算量・複雑性の観点から難しい.
    そのため, 機能的関係が深いセンサ・アクチュエータをグルーピングし, それらのグループに対してそれぞれ制御器やネットワークを構築することが有効である.
    本研究では, センサ・アクチュエータ同士の機能的な接続と空間的な接続をグラフ構造に落とし込み, 自動的にグルーピングする手法を開発する.
    多数の冗長な筋を持つ筋骨格ヒューマノイドの筋分割を例に取り, これまで幾何モデルから人間が行ってきた前腕・上腕・肩甲骨・首周辺と言った部位ごとの筋分割を, 幾何モデルなしに筋同士の機能的関係性と神経接続に見る空間的配置を考慮し, 自動的に行う.
    筋骨格ヒューマノイドMusashi, Kengoroの筋分割に本研究を適用し, その有効性を示す.
  }%
\end{abstract}

\section{INTRODUCTION}\label{sec:introduction}
\begin{figure}[t]
  \centering
  \includegraphics[width=0.7\columnwidth]{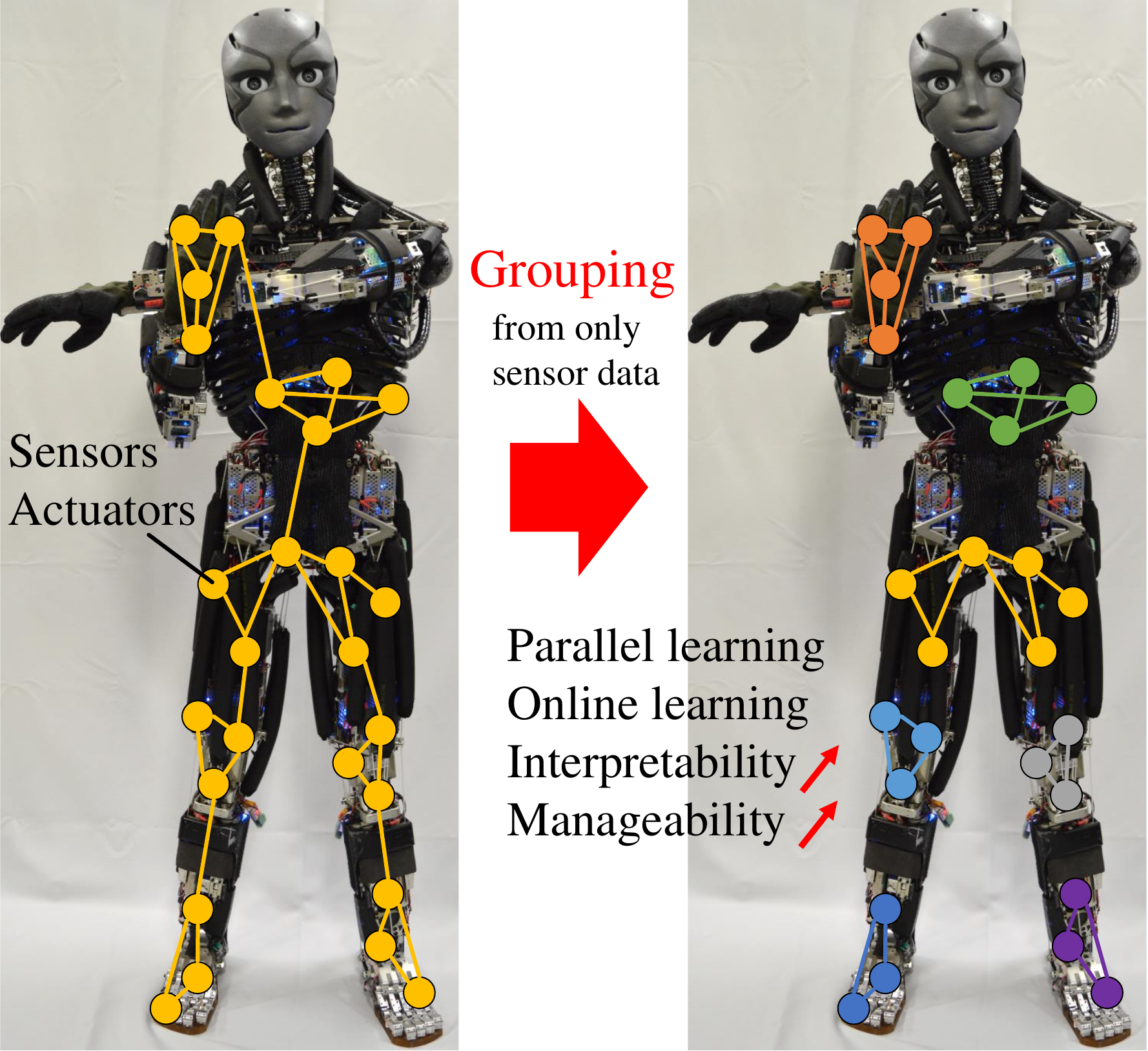}
  \vspace{-1.0ex}
  \caption{The concept of this study.}
  \label{figure:concept}
  \vspace{-3.0ex}
\end{figure}

\switchlanguage%
{%
  For a robot with redundant sensors and actuators distributed throughout its body, it is computationally difficult to construct a single controller or neural network using all of them.
  Hence, the reinforcement learning \cite{duan2016benchmarking} that integrates all sensors and actuators and the online learning \cite{kawaharazuka2020autoencoder} that is prone to overfitting are still difficult to conduct.
  Also, sensors and actuators distributed throughout the body \cite{mittendorfer2011tactile} are characterized by the fact that the functions of sensors and actuators are easily divided into different regions of the body such as the fingers, hip, and feet.
  Therefore, while some tasks require the use of sensors and actuators of the whole body at the same time, extracting functionally related sensors and actuators, grouping them, and constructing a controller or a network for each of these groups is often effective (\figref{figure:concept}).
  The grouping improves the interpretability and manageability, and also enables online learning of the individual networks in parallel.
  In this study, the functional and spatial connections among sensors and actuators are embedded into a graph structure and the method for automatic grouping is developed (spatial connection is used as a support for functional connection).
  We apply this method to musculoskeletal humanoids with redundant muscles and verify its effectiveness.

  The musculoskeletal humanoid \cite{asano2016kengoro, kawaharazuka2019musashi}, mimicking not only the structure of the human body, but also the muscle actuator, has many redundant muscles as with humans.
  This redundancy is an important element and enables the robust continuous movement even when one muscle breaks \cite{kawamura2016jointspace} and variable stiffness control with nonlinear elastic elements \cite{kawaharazuka2019longtime}.
  At the same time, it is very difficult to manage and move a large number of redundant muscles distributed throughout the body by a single controller or a single neural network in terms of computational cost and complexity.

  Therefore, previous control and state estimation methods have divided the muscles into regions with weak relationships and constructed controllers and neural networks for each of them.
  In \cite{kawaharazuka2019longtime, kawaharazuka2020autoencoder}, a neural network is constructed for each group of actuators and sensors, and the control, state estimation, and simulation are performed for each group.
  With the appropriate grouping, the number of actuators and sensors involved is limited to a small number, and the online learning is successfully performed with a small amount of computation.
  A torque controller has been constructed in the same way in \cite{kawamura2016jointspace}.
  In the case of existing polyarticular muscles, a rough guide for muscle grouping is presented in \cite{kawaharazuka2018estimator} and a method to perform the accurate joint angle estimation based on the muscle grouping is discussed.
  Most controllers are applied only to a part of the arm, such as \cite{jantsch2011scalable, kawaharazuka2020dynamics}, and their applications to the whole body have not been discussed much so far.
}%
{%
  冗長なセンサ・アクチュエータが全身に分布するロボットにおいて, それら全てを用いて一つの制御器やニューラルネットワークを構築することは計算量・複雑性の観点から難しい.
  ゆえに, 全センサ・アクチュエータを統合して使用する強化学習\cite{duan2016benchmarking}や過学習を起こしやすいオンライン学習\cite{kawaharazuka2020autoencoder}は現在でも難しい.
  同時に, 全身に分布するセンサ・アクチュエータは, 指や腰, 足等の部位ごとに機能が分かれやすいという特徴もある.
  そのため, 全身のセンサ・アクチュエータを同時に用いる必要のあるタスクもある一方, 関係が深いセンサ・アクチュエータをグルーピングし, それらのグループに対してそれぞれ制御器やネットワークを構築することが有効な場合も多い(\figref{figure:concept}).
  グルーピングを施すことで, 解釈性・扱いやすいさが向上し, 並列でそれぞれの学習器をオンライン学習させることも可能となる.
  本研究では, センサ・アクチュエータ同士の機能的な接続と空間的な接続をグラフ構造に落とし込み, 自動的にグルーピングする手法を開発する.
  本手法を冗長な筋を持つ筋骨格ヒューマノイドに適用しその有効性を示す.

  人体の構造だけでなく駆動方法さえも模倣した筋骨格ヒューマノイドには, 人間と同様な多数の冗長な筋が存在する.
  この冗長性は重要な要素であり, 筋が一本切れても動き続けるロバスト性や\cite{kawamura2016jointspace}, 非線形弾性要素を併用することによる可変剛性制御\cite{kawaharazuka2019longtime}を可能とする.
  同時に, 全身に分布する多数の冗長な筋を一つの制御器や一つのニューラルネットワークによって管理し動かすことは, 計算量や複雑性の観点から非常に困難である.

  そのため, これまでの制御手法・状態推定手法では, 関係性が希薄な部位ごとに筋群を適切に分割し, それぞれに対して制御器やニューラルネットワークを構築してきた.
  \cite{kawaharazuka2019longtime, kawaharazuka2020autoencoder}では, 適切にグルーピングした筋アクチュエータ・センサ群それぞれにおいてニューラルネットワークを構築し, 制御や状態推定・シミュレーションを実行している.
  適切なグルーピングによって関わるアクチュエータ・センサを少数に限ることで, 小さな計算量でオンライン学習を行うことにも成功している.
  \cite{kawamura2016jointspace}でも同様にトルクコントローラを構築している.
  \cite{kawaharazuka2018estimator}では多関節筋がある場合における筋分割の目安を提示し, これを元に正確な関節角度推定を実行する方法について考察している.
  また, 大抵の制御器は全身ではなく腕の一部\cite{jantsch2011scalable, kawaharazuka2020dynamics}等のみに適用されており, 全身に関する適用についてはあまり議論されていない.
}%

\begin{figure*}[t]
  \centering
  \includegraphics[width=1.8\columnwidth]{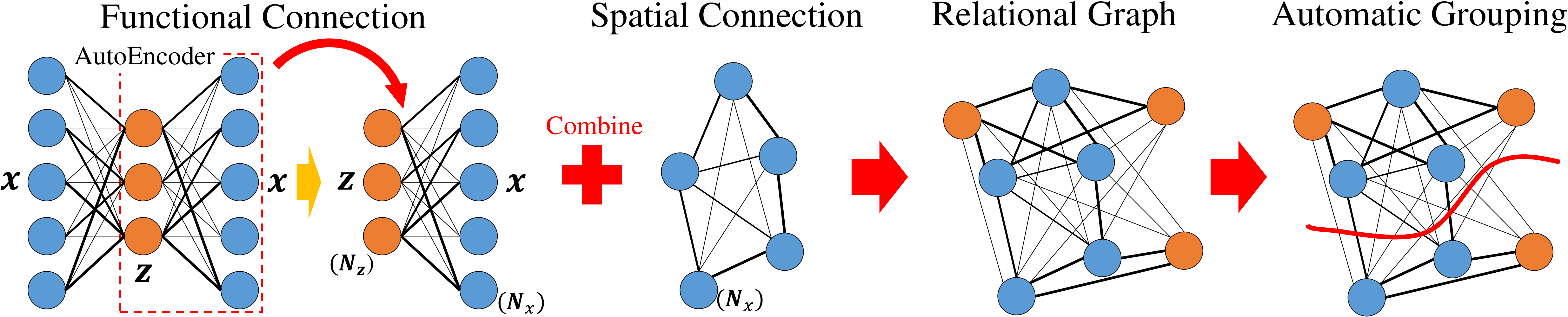}
  \vspace{-1.0ex}
  \caption{The overall flow of automatic grouping using a relational graph with functional and spatial connections.}
  \label{figure:connection-graph}
  \vspace{-3.0ex}
\end{figure*}

\switchlanguage%
{%
  There are two types of information to help in grouping a large number of redundant muscles located throughout the body: functional and spatial connections.
  The functional connection is that the muscles are functionally related to each other due to their redundancy, which indicates the strength of their correlation (e.g. the relationship between the agonist and the antagonist muscles is strong).
  The spatial connection is a measure of the spatial proximity of muscles that comes from neural connections (e.g. the spatial connection between the leg and arm muscles is weak).
  By embedding these two pieces of information into a graph structure, the appropriate muscle grouping is automatically performed by conducting graph partitioning, while each one of them alone is not enough.
  This method also enables the robot to automatically build a highly interpretable and easy-to-use controller with high accuracy for each group while reducing the computational cost and complexity, since only necessary edges and nodes remain.
  Since these groupings have been considered by human beings in the past, the problem of automatically determining the groupings from sensor data alone is new.
  In addition, although a method using EMG signals of the human body \cite{niiyama2010emg} is possible only for the musculoskeletal system, we do not implement an algorithm specific to the musculoskeletal system in this research so that it can be applied to various robots in the same way.
  Although methods using information distance \cite{olsson2004grouping} have been proposed so far, these methods provide not explicit grouping of sensors and actuators but only correlational relationships among them.

  The contributions of this paper are as follows.
  \begin{itemize}
    \item Embedding of functional and spatial connections of sensors and actuators into a graph structure
    \item Development of a grouping method of sensors and actuators using the relational graph
    \item Application of the proposed method to muscle grouping of musculoskeletal humanoids
  \end{itemize}
}%
{%
  全身に配置された多数の冗長な筋群をグルーピングするときに助けとなる情報には, 機能的接続と空間的接続がある.
  機能的接続とは, 筋は冗長であるがゆえにそれぞれが機能的に関係しあっており, その相関の強さを表す(e.g. 主動筋と拮抗筋の関係性は強い).
  空間的接続とは, 神経的な接続から来る筋同士の空間的な近さを表す(e.g. 足の筋と腕の筋の空間的な接続は弱い).
  この2つの情報をグラフ構造に埋め込み, グラフ分割を行うことで, 適切な筋グルーピングを自動的に行う.
  本手法を適用することにより, ロボットは自身のランダムな動きから機能的接続を見出し, これと空間的接続を合わせることで適切な筋グルーピングを行うことができる.
  また, 計算量や複雑性を下げつつも正確性を保った解釈性が高く扱いやすい制御器をグルーピングごとに自動的に作り上げていくことが可能となる.

  これまで, これらグルーピングは人間が考えて行うものであったため, それをセンサデータのみから自動的に決定する本研究の問題設定は今までになく新しい.
  また, 筋骨格系に限っては人体のEMGを模倣する方法も考えられるが, 本研究では様々なロボットに同様に適用できるよう, 筋骨格系にspecificなアルゴリズム実装は行わない.
  本手法を筋骨格ヒューマノイドMusashi, Kengoroに適用し, これまで幾何モデルから人間が行ってきた筋グルーピングとの比較, グルーピング後の学習性能の比較を行う.

  本研究の構成は以下のようになっている.
  \secref{sec:proposed}では, 一般化した形で機能的な・空間的接続のグラフ構造への埋め込み, この関係グラフを使ったgrouping方法について提案する.
  \secref{sec:muscle-grouping}では, 筋骨格ヒューマノイドの筋構造を例にその機能的・空間的関係性について議論する.
  \secref{sec:experiments}では, 筋骨格ヒューマノイドMusashiとKengoroに対して本研究を適用し, その有効性を検証する.
  最後に, 議論と結論を述べる.
}%

\section{Automatic Grouping of Redundant Sensors and Actuators Using Functional and Spatial Connections} \label{sec:proposed}
\switchlanguage%
{%
  In this study, the value of redundant sensors and actuators is expressed as $\bm{x}$ (its dimension is $N_{x}$).
  All the obtained data is normalized in order to eliminate the scale difference of each value.
  The connection between $\bm{x}$ and a latent variable $\bm{z}$, which will be explained below, is given as weighted undirected edges and automatically divided by a randomized selection algorithm.
  The entire flow is shown in \figref{figure:connection-graph}.
}%
{%
  本研究では, ある冗長なセンサやアクチュエータの値を$\bm{x}$とし, その次元を$N_{x}$とする.
  ただし, それぞれのデータ間におけるスケールの違いを無くすため, 得られた全データを使って正規化を施すこととする.
  この$\bm{x}$, そして以降で説明する潜在変数$\bm{z}$を合わせた変数群に重み付きの無向辺を張り, 乱択アルゴリズムによって自動分割する方法について説明する.
  全体のフローを\figref{figure:connection-graph}に示す.

}%

\subsection{Relational Graph with Functional Connections} \label{subsec:functional-connection}
\switchlanguage%
{%
  Functional connection expresses the relationship among sensors and actuators through latent variables.
  Since $\bm{x}$ is redundant, there is some latent variable $\bm{z}$, whose dimension is $N_{z}$ ($N_{z}<N_{x}$) and the relation among $\bm{x}$ is represented by $\bm{z}$.
  That is, the function of $\bm{x}$ can be represented by $\bm{z}$.

  This functional connection can be trained by using AutoEncoder \cite{hinton2006reducing}.
  By training an AutoEncoder in which the input is $\bm{x}$, the middle layer as the bottleneck is $\bm{z}$, and the output is $\bm{x}$, the functional connection among $\bm{x}$ can be calculated via $\bm{z}$.
  If the AutoEncoder is three-layered, the weight matrix $W$ (${N_{z}}\times{N_{x}}$) between the second and third layers simply expresses the weighted undirected edges between the nodes of $\bm{z}$ and $\bm{x}$ in the graph structure.
  This is a bipartite graph with no edges among $\bm{z}$ nodes and among $\bm{x}$ nodes.
  Even if the AutoEncoder is not designed to extract $W$ directly, as with $2N_{AE}+1$ layers ($N_{AE}$ is a constant), it is still possible to calculate $W$ with the form of $W=\prod^{2N_{AE}}_{k=N_{AE}+1} W_{k}$, where $W_{k}$ is the weight matrix between the $k$-th and $(k+1)$-th layers.
  For the connections in $W$, the larger the value, the stronger the functional connection and the more likely the corresponding nodes are divided into the same group.
}%
{%
  ここでいう機能的な接続とは, $\bm{x}$が冗長であるがゆえに, 何らかの潜在変数$\bm{z}$(次元数は$N_{z}$, $N_{z}<N_{x}$)が存在し, $\bm{x}$の間の関係は$\bm{z}$によって表されることを指す.
  つまり, $\bm{x}$の機能が$\bm{z}$によって表現可能である.

  この機能的な接続を学習させることはAutoEncoder \cite{hinton2006reducing}を用いることで可能である.
  入力を$\bm{x}$, ボトルネックとなる中間層を$\bm{z}$, 出力を$\bm{x}$とするようなAntoEncoderを学習させることで, $\bm{x}$の機能的な接続を$\bm{z}$を介して計算することができる.
  もしこのAutoEncoderが3層である場合は, 2層と3層の間の重み$W$ (${N_{z}}\times{N_{x}}$)が, グラフ構造に置いて$\bm{z}$と$\bm{x}$をノードとしたときの, ノード間の重み付き無向辺となる.
  これは, $\bm{z}$同士, $\bm{x}$同士の間には辺がないような二部グラフとなっている.
  もしAutoEncoderが5層のように直接$W$を取り出せないような構造であったとしても, 3層目と4層目の間の重みを$W_{1}$, 4層と5層の間の重みを$W_{2}$として, $W=W_{1}W_{2}$という形で$W$を計算することが可能である.
  $W$の中の接続で, 値が大きなものほどより機能的な結びつきが強く, 同じグルーピングに分けられやすい.
}%

\subsection{Relational Graph with Spatial Connections} \label{subsec:spatial-connection}
\switchlanguage%
{%
  Spatial connection here is a constraint among $\bm{x}$, such as spatial closeness in the body, less delay in neural connections, or being connected to the same circuit.
  We embed this spatial connection into the graph as a weighted edge between the two coordinates of the sensors $\bm{x}$ in order to represent the fact that the closer they are spatially, the more likely they are to be divided into the same group.
  In order to be consistent with the evaluation that higher edge weights in functional connections tend to result in the same group, it is necessary to set the values so that edge weights become higher as spatial connections become closer.
  Assuming that the spatial distance is represented as $d$, we embed the edge weights as $-d$ in this study.
}%
{%
  ここでいう空間的な接続とは, $\bm{x}$間について, 空間的な近さや, 同じ回路に接続していなどの制約を表す.
  空間的に近い値ほど, 同じグルーピングに分けられやすいことを表現するために, この空間的な接続を$\bm{x}$同士の間の重み付きの辺としてグラフ内に埋め込む.
  このとき, 機能的接続における辺の重みが大きいほど同じグルーピングになりやすいという評価と一致するよう, 空間的接続が近いほど辺の重みが大きくなるように値を設定する必要がある.
  空間的な距離が$d$と表せたとすると, 本研究では辺の重みを$-d$として埋め込む.
}%

\subsection{Automatic Grouping Method Using Relational Graph}
\begin{figure}[t]
  \centering
  \includegraphics[width=0.8\columnwidth]{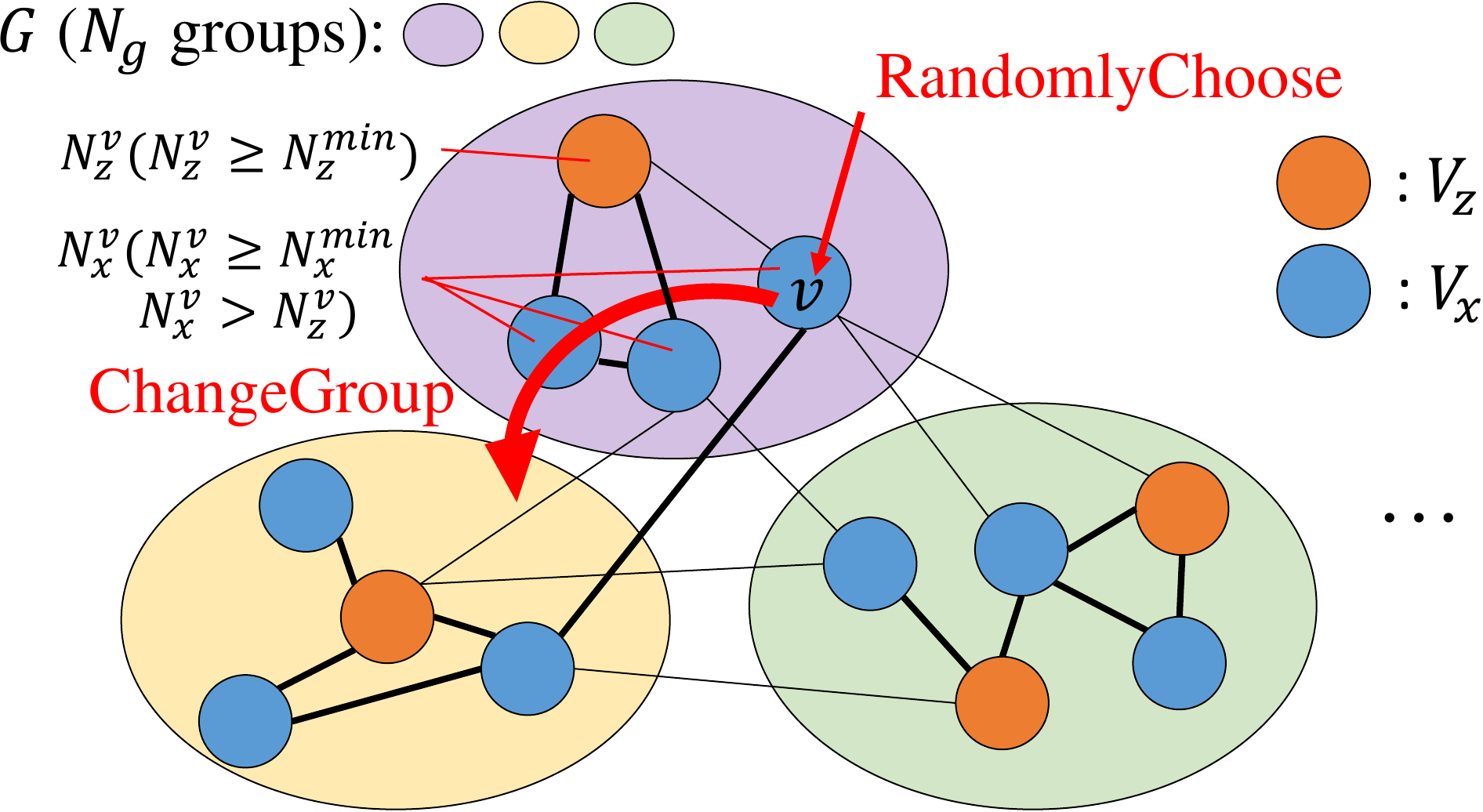}
  \vspace{-1.0ex}
  \caption{Conceptual diagram of automatic grouping.}
  \label{figure:grouping}
  \vspace{-1.0ex}
\end{figure}

\switchlanguage%
{%
  The relational graph in \figref{figure:connection-graph} is constructed by combining functional and spatial connections.
  The grouping in this study corresponds to cutting its edges with small weights, i.e., the edges with weak relations, and grouping the vertices with edges of high weights, i.e., the vertices with strong relations.
  The two most promising methods for solving this problem are the minimum cut algorithm and the minimum spanning tree algorithm.
  For the minimum cut, we can use algorithms such as \cite{nagamochi2002mincut} that do not require definitions of source and sink.
  Also, the minimum spanning tree can be applied to multiple groups by merging vertices with Kruskal method \cite{kruskal1956spanning}, until the number of groups becomes the desired number.
  However, when these algorithms are applied to relational graphs containing functional connections in this study, extreme solutions such as grouping into one vertex and $N_{x}-1$ vertices are almost always optimal, and we are unable to create a balanced grouping such that each group contains roughly the same number of vertices.
  This is due to the fact that the weights obtained by AutoEncoder are not exactly zero for edges whose weights should be zero, but each edge has some weight and cannot be divided clearly.

  In this study, we apply an algorithm that takes into account the constraints on the number of vertices among groups as follows (\figref{figure:grouping}).
  For each node in the obtained relational graph, we assign an appropriate group label.
  In this study, the number of groups to be divided into is set to $N_{g}$, and as a constraint, the minimum number of vertices of $\{\bm{x}, \bm{z}\}$ in a group is set to $N^{min}_{\{x, z\}}$.
  The notations for this algorithm are shown in \tabref{table:notations}, and the pseudo code is shown in \algoref{algorithm:grouping}.
 }%
{%
  本研究におけるグルーピングは, 辺の重みが小さい, つまり関係が希薄な辺を切って, 辺の重みが大きい, つまり関係が濃い頂点同士をまとめていくことに相当する.
  この問題を解く最も有力な方法として, 最小カットのアルゴリズム, 最小全域木のアルゴリズムが考えられる.
  最小カットについては, sourceとsinkの定義が必要ない\cite{nagamochi2002mincut}等のアルゴリズムを用いることが可能である.
  また, 最小全域木については, Kruskal法\cite{kruskal1956spanning}において所望のグルーピング数になるまで頂点をマージすることで複数グルーピングにも適用可能である.
  しかし, これらのアルゴリズムを本研究の機能的接続のみを含む関係グラフに適用したところ, 頂点1個と頂点$N_{x}-1$個等の極端な解が最適回となることがほどんどであり, それぞれのグルーピングが大体同じだけの頂点を含むようなバランスの良いグルーピングを作ることができなかった.
  これは, AutoEncoderによって得られた重みについて, 関係がない辺の重みがきっちり0になるわけではなく, それぞれある程度の重みを持っており, 綺麗に分割ができないことが原因であると考えられる.

  本研究では, 以降に示すようなグルーピング間の頂点数についての制約等を考慮したアルゴリズムを適用する.
  得られたグラフに対して, それぞれのノードについてグループのラベルを振り分ける.
  本研究では, 分割したいグルーピング数を$N_{g}$とし, 一つのグループに含まれる最小の$\bm{z}$の頂点数を$N^{min}_{z}$, 最小の$\bm{x}$の頂点数を$N^{min}_{x}$として制約を設定する.
  また, グラフの全頂点の集合を$V$, $\bm{x}$に含まれる頂点の集合を$V_{x}$, $\bm{z}$に含まれる頂点の集合を$V_{z}$, 全グループの集合を$G$, 全エッジを$E$, 頂点$v \in V$から出るエッジを$E_{v}$, 頂点$v \in V$から集合$g \in G$に含まれる頂点への辺の集合を$E^{g}_{v}$, 頂点$v$が含まれる集合$g \in G$における$\bm{x}$に含まれる頂点の数を$N^{v}_{x}$, 頂点$v$が含まれる集合$g \in G$における$\bm{z}$に含まれる頂点の数を$N^{v}_{z}$とする.
  擬似コードを\algoref{algorithm:grouping}に示す.
}%

\begin{table}[t]
  \centering
  \caption{Notations in this paper}
  \vspace{-1.0ex}
  \label{table:notations}
  \begin{tabular}{|c|c|}
    \hline
    Notation & Definition \\
    \hline \hline
    $V$ & set of all vertices of the relational graph\\\hline
    $V_{\{x, z\}}$ & set of vertices included in $\{\bm{x}, \bm{z}\}$\\\hline
    $G$ & set of all groups\\\hline
    $E$ & set of all edges of the relational graph\\\hline
    $E_{v}$ & set of edges from the vertex $v \in V$\\\hline
    $E^{g}_{v}$ & set of edges from the vertex $v \in V$ to the vertices in $g \in G$\\\hline
    $N^{v}_{\{x, z\}}$ & the number of vertices of $\{\bm{x}, \bm{z}\}$ included in\\
    & the group $g \in G$ to which the vertex $v \in V$ belongs\\\hline
  \end{tabular}
  \vspace{-3.0ex}
\end{table}

\begin{algorithm}[t]
\small
  \caption{Automatic grouping method}
  \label{algorithm:grouping}
  \begin{algorithmic}[1]
    \Function{Grouping}{}
      \State $\textrm{InitializeGroup}(V)$
      \State $n_{iter} \gets 0$
      \While{$n_{iter} < N_{iter}$}
        \State $v \gets \textrm{RandomlyChoose}(V)$
        \If{$v \in V_{x}$ \AND ($N^{v}_{x} \leq N^{min}_{x}$ \OR $N^{v}_{x} \leq N^{v}_{z}+1$)}
          \State \textbf{Continue}
        \EndIf
        \If{$v \in V_{z}$ \AND $N^{v}_{z} \leq N^{min}_{z}$}
          \State \textbf{Continue}
        \EndIf
        \State $\bm{S} \gets \textrm{InitializeEval}(G)$
        \For{$g \in G$}
          \State $s \gets \textrm{CalcEval}(E^{g}_{v})$
          \State $\textrm{SetEval}(\bm{S}, g, s)$
        \EndFor
        \State $\textrm{SortByEval}(\bm{S})$
        \State $n \gets n_{iter}/N_{iter}$
        \State $\textrm{ChangeGroup}(v, \bm{S}, n)$
        \State $n_{iter} \gets n_{iter}+1$
      \EndWhile
    \EndFunction
  \end{algorithmic}
\normalsize
\end{algorithm}

\switchlanguage%
{%
  Here, $N_{iter}$ is the number of iterations of the randomized selection algorithm.
  Also, $\textrm{InitializeGroup}(V)$ is an operation that randomly assigns a group label to each vertex of $V$.
  $\textrm{RandomlyChoose}(V)$ is an operation that randomly selects one vertex in $V$.
  $\textrm{InitializeEval}(G)$ is a function that initializes and returns a vector of evaluation values for each group $\bm{S}$.
  $\textrm{SetEval}(\bm{S}, g, s)$ is an operation that sets the value of $\bm{S}$ to $s$ for the group $g$.
  $\textrm{CalcEval}(E^{g}_{v})$ is a function that returns the evaluation value obtained from the weights of the edges representing the functional and spatial connections in $E^{g}_{v}$.
  $\textrm{SortByEval}(\bm{S})$ is an operation that rearranges $\bm{S}$ in descending order.
  $\textrm{ChangeGroup}(v, \bm{S}, n)$ is an operation to change the group, which a vertex $v$ belongs to, to the group with the highest evaluation value of $\bm{S}$ according to $n$ ($n$ is described later).

  The algorithm is a simple algorithm that first randomly initializes the group to which each vertex belongs, chooses a random vertex among them, and then changes the vertex from the current group to one of the groups (including the current group) based on the evaluation function.
  Here, Line 6--10 is a condition for satisfying the constraint on the minimum number of vertices of $\bm{x}$ and $\bm{z}$ in a group ($N^{v}_{\{x, z\}} \geq N^{min}_{\{x, z\}}$) and the constraint that the number of vertices of the latent variable $\bm{z}$ is less than the number of vertices of $\bm{x}$ ($N^{v}_{x} > N^{v}_{z}$).
  Line 12--17 computes the evaluation value of $v$ when $v$ is changed to belong to each group in $G$ and calculates to which group $v$ should belong to obtain the highest value.
  In Line 14, we calculate and sum up the evaluation values of each functional and spatial connections as follows,
  \begin{align}
    \textrm{CalcEval}(E^{g}_{v}) = 1/N_{func}\Sigma{w_{func}}+\alpha\Sigma{w_{spac}} \label{eq:eval}
  \end{align}
  where $w_{\{func, spac\}}$ is each weight of edges of \{functional, spatial\} connections in $E^{g}_{v}$, $N_{func}$ is the number of functional edges in $E^{g}_{v}$, and $\alpha$ is a constant weight for the evaluation values.
  As mentioned earlier, functional connections are trained by AutoEncoder and therefore do not become zero even if the vertices are not related to each other.
  Therefore, if we take the sum of the weights of the functionally connected edges in $E^{g}_{v}$, the number of edges in the group with the largest number of vertices will increase, and so the average of the weights is used as the evaluation value.
  Also, since the weights of the spatially connected edges are negative as described above and the evaluation value decreases as the number of edges increases, there is no need to use the average, and so the sum of the weights is used as the evaluation value.
  It is possible to express the weights of spatially connected edges as positive values such as $1/d$ and treat them in a unified manner, but this did not work well in this study.
  When only functional or spatial connections are used in \secref{sec:experiments}, either one of the evaluation values is calculated.
  Finally, in Line 19, when we change the group of vertex $v$, we change the behavior of the grouping by $n=n_{iter}/N_{iter}$.
  If we always select the group with the highest evaluation value, the algorithm would immediately fall into a localized solution, and so, as in the annealing method, the group with the highest evaluation value is selected as $n$ nears 1, and the group is selected randomly as $n$ nears 0.
  In other words, the group with the highest evaluation value is selected with $n$ probability and the random group is selected with $1-n$ probability.

  In this study, we set $N^{min}_{x}=2$, $N^{min}_{z}=1$, $N_{iter}=30000$, and $\alpha=10$.
  Since the value of the first term on the right-hand side of \equref{eq:eval} is large and the value of its second term is negative near 0, we can balance the two evaluation values by setting $\alpha$ to be $\alpha>1$.
}%
{%
  ここで, $N_{iter}$は乱択アルゴリズムのイテレーション回数を表す.
  また, $\textrm{InitializeGroup}(V)$は$V$のそれぞれの頂点にランダムにグループのラベルを振り分ける操作である.
  $\textrm{RandomlyChoose}(V)$は$V$に含まれる頂点をランダムに一つ選ぶ操作である.
  $\textrm{InitializeEval}(G)$はそれぞれのグルーピングに対する評価値のベクトルを初期化して返す関数である.
  $\textrm{SetEval}(\bm{S}, g, s)$はグルーピング$g$に関する$\bm{S}$の値を$s$にセットする操作である.
  $\textrm{CalcEval}(E^{g}_{v})$は$E^{g}_{v}$に含まれる機能的・または空間的な接続を表す辺の重みから得られる評価値を返す関数である.
  $\textrm{SortByEval}(\bm{S})$は$\bm{S}$を評価値の降順に並べ変える操作である.
  $\textrm{ChangeGroup}(v, \bm{S}, n)$は頂点$v$の属するグループを$\bm{S}$の評価値が最も高いグループに変更する操作である($n$については後に述べる).

  本アルゴリズムは, まずそれぞれの頂点が属するグループを適当に選択・初期化し, その中からランダムな頂点を一つ選び, その頂点を現在のグループから評価関数をもとにいずれかのグループ(現在のグループを含む)に変更するという単純なアルゴリズムである.
  このとき, Line 6--10は一つのグループに含まれる$\bm{x}$, $\bm{z}$の頂点の最小値制約, また, 潜在変数$\bm{z}$の頂点の方が$\bm{x}$の頂点数よりも少ないという制約を満たすための条件である.
  また, Line 12--17は$v$を$G$に含まれるそれぞれのグループに属するように変更した場合の評価値を計算し, どのグループに属することで評価値が最も高くなるかを計算する.
  Line 14では, 以下のように機能的接続と空間的接続それぞれについて評価値を計算し合計する.
  \begin{align}
    \textrm{CalcEval}(E^{g}_{v}) = 1/N_{Func}\Sigma{w^{Func}}+\alpha\Sigma{w^{Spac}} \label{eq:eval}
  \end{align}
  ここで, $w^{\{Func, Spac\}}$は$E^{g}_{v}$の辺における機能的または空間的接続に関する辺の重みを表し, $N_{Func}$は$E^{g}_{v}$に含まれる機能的接続辺の数, $\alpha$は評価値の重みを表す定数である.
  機能的接続は先に述べたように, AutoEncoderで学習されるため互いに関係がなくても重みが0になるわけではない.
  よって, $E^{g}_{v}$に含まれる機能的接続辺の重みの合計を評価値とすると, 最も頂点数が多いグループへの辺が増えそちらに引っ張られやすくなってしまうため, 重みの平均を評価値として用いる.
  また, 空間的接続辺の重みは先に述べたように負であり, 辺が増えるほど評価値が下がるため, 平均を用いる必要はなく, 重みの合計を評価値として用いる.
  空間的接続辺の重みを$1/d$のように正の値で表し統一的に扱うことも可能であるが, 本研究では上手く動作しなかった.
  \secref{sec:experiments}において機能的接続, または空間的接続のみを利用する場合は, 一方の評価値のみ計算することになる.
  最後にLine 19だが, ここで頂点$v$のグルーピングを変更する際, $n=n_{iter}/N_{iter}$によってグループ変更の挙動を変化させる.
  常に最も評価値が高いグループに変更させてしまうと, すぐに局所解に陥ってしまうため, 焼きなまし法のように, $n$が1に近づくほど最大の評価値のグループが選ばれ, $n$が0に近いほどランダムにグループが選ばれるようにする.
  つまり, $n$の確率で最大評価値のグループを選び, $1-n$の確率でランダムにグループを選ぶ.

  本研究では, $N^{min}_{x}=2$, $N^{min}_{z}=1$, $N_{iter}=30000$, $\alpha=10$とする.
  \equref{eq:eval}右辺第一項の値は大きく, \equref{eq:eval}の右辺第二項の値は0に近い負の値となるため, $\alpha$は$>1$となるように設定することで二つの評価値をバランスさせることができる.
}%

\section{Muscle Grouping for Musculoskeletal Humanoids} \label{sec:muscle-grouping}
\switchlanguage%
{%
  In this study, the method proposed in \secref{sec:proposed} is applied to muscle grouping for musculoskeletal humanoids.
  The structure of musculoskeletal humanoids and the functional and spatial connections in muscles are explained, and the effectiveness of the method is verified by experiments in \secref{sec:experiments}.
}%
{%
  本研究では, \secref{sec:proposed}において提案した手法を筋骨格ヒューマノイドにおける筋分割に適用する.
  筋骨格ヒューマノイドの構造, 筋における機能的・空間的接続とは何かについて説明し, \secref{sec:experiments}においてその有効性を実験により示す.
}%

\begin{figure}[t]
  \centering
  \includegraphics[width=0.7\columnwidth]{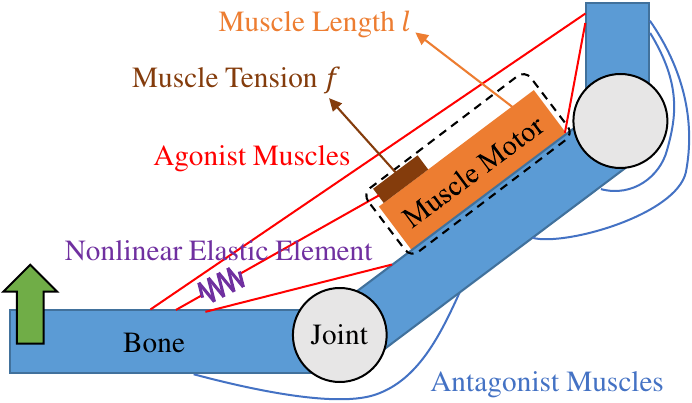}
  \vspace{-1.0ex}
  \caption{The basic musculoskeletal structure.}
  \label{figure:musculoskeletal-structure}
  \vspace{-3.0ex}
\end{figure}

\subsection{The Basic Structure of the Musculoskeletal Humanoid} \label{subsec:musculoskeletal-structure}
\switchlanguage%
{%
  The basic musculoskeletal structure is shown in \figref{figure:musculoskeletal-structure}.
  Redundant muscles are antagonistically arranged around the joints.
  There are not only monoarticular muscles acting on a single joint but also polyarticular muscles acting on multiple joints at the same time.
  The redundancy of these muscles enables the joint to move even if one muscle is broken, and provides for variable stiffness control with nonlinear elastic elements.
  For each muscle, muscle length $l$ and muscle tension $f$ can be measured from the encoder and loadcell, respectively.
  The joint angle $\bm{\theta}$ is usually difficult to measure due to ball joints or the complex scapula (although some robots can measure it like in \cite{kawaharazuka2019musashi}).

  In this study, we discuss how to group muscles without prior knowledge of the arrangement of joints and muscles.
}%
{%
  基本的な筋骨格構造の構成を\figref{figure:musculoskeletal-structure}に示す.
  冗長な筋が関節の周りに拮抗して配置されている.
  一つの関節に作用する単関節だけでなく, 複数の関節にまたがって作用するような多関節筋が同時に含まれる.
  この筋が冗長なことにより, 一本の筋が切れても関節を動かすことができる, また, 非線形弾性要素と合わせて可変剛性制御が可能となる.
  筋は主に摩擦に強い合成繊維であるDyneemaによって構成されており, それぞれの筋についてエンコーダから筋長$l$・ロードセルから筋張力$f$・温度センサから筋温度$c$が測定できる.
  関節角度$\bm{\theta}$は球関節や複雑な肩甲骨ゆえに測定できない場合が多い(一部のロボットで測定することが可能である\cite{kawaharazuka2019musashi}).

  本研究では関節や筋の配置に関する事前知識がない状態において, 如何に筋をグルーピングするかについて議論する.
}%

\subsection{Functional Connections of Muscles} \label{subsec:functional-muscle}
\switchlanguage%
{%
  This method is applied to the musculoskeletal structure with the muscle length $\bm{l}$ as $\bm{x}$ in \secref{sec:proposed}.
  The muscle length $\bm{l}$ is redundant, and when a joint moves in a certain direction, there are always more than two muscles around the joint: the agonist muscle, which carries the movement, and the antagonist muscle, which prevents the movement.
  Therefore, the muscle length $\bm{l}$ can be represented by the latent variable $\bm{z}$ as seen in \secref{subsec:functional-connection} (if $\bm{x}$ is $\bm{l}$, then $\bm{z}$ can be defined as $\bm{\theta}$, but we handle it as a latent variable $\bm{z}$ in this study, because we do not have any prior knowledge about joints and we cannot obtain the joint angles on the actual robot).
  As explained above, there are polyarticular muscles, and therefore there are cases in which muscles span more than one group.
  In this study, each muscle is grouped into one group, but it is also possible to group the muscles into multiple groups by observing the functional connection of each muscle after the grouping (this is one of our future works).
  We obtain the data of $\bm{l}$ from the random movements of muscles and use it to train the AutoEncoder.
}%
{%
  筋骨格構造における筋長$\bm{l}$を\secref{sec:proposed}における$\bm{x}$として本手法を適用する.
  この筋長$\bm{l}$は冗長な構造をしており, ある方向に関節を動かす時, その関節周りに必ず動きを担う主動筋と動きを妨げる拮抗筋が存在する.
  そのため, これらの筋長$\bm{l}$は\secref{subsec:functional-connection}に見るように潜在変数$\bm{z}$によって表現可能である($\bm{l}$を$\bm{x}$とする場合は$\bm{\theta}$を$\bm{z}$と同様に用いることが可能であるが, 本研究は関節に関する前提知識はなく, 実機に置いても基本的に関節角度は得られないため, 本研究では潜在変数$\bm{z}$として扱う).
  説明したように筋には多関節筋があり, そのため複数のグルーピングに筋がまたがる場合も存在する.
  本研究ではこれをいずれか一方のグループに所属させるが, グルーピング後にそれぞれの筋について機能的接続を見て複数グループに所属させることも可能である.
  筋のランダムな動きから$\bm{l}$のデータを取得し, これを使ってAutoEncoderを学習させる.
}%

\begin{figure}[t]
  \centering
  \includegraphics[width=0.8\columnwidth]{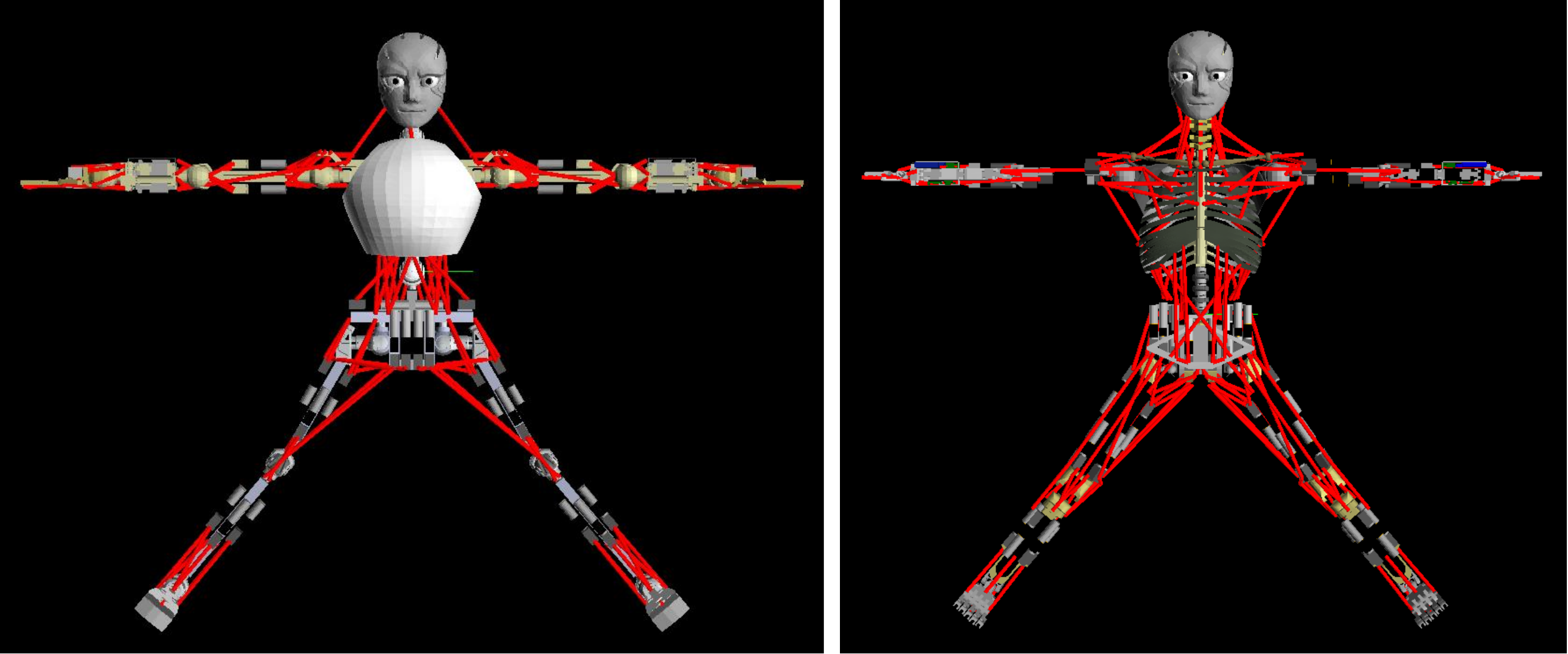}
  \vspace{-1.0ex}
  \caption{The posture of Musashi (left) and Kengoro (right) when calculating spatial connections in this study.}
  \label{figure:spatial-muscle}
  \vspace{-3.0ex}
\end{figure}

\subsection{Spatial Connections of Muscles} \label{subsec:spatial-muscle}
\switchlanguage%
{%
  In humans, the closer the muscles are to each other spatially, the stronger the neural connections are, and this concept is incorporated as spatial connections.
  Since there is no prior knowledge of the muscle arrangement, we should introduce an index expressing whether the communication connection has less delay in terms of the circuitry (i.e., they are spatially close to each other), but it is difficult in the current condition.
  Therefore, we do not use the information on the muscle arrangement directly, but only use a matrix of how far the center of each muscle path in the geometric model is from each other.
  We use the spatial distance among the muscles with the arms and legs spread, as an approximate spatial connection distance, as in \figref{figure:spatial-muscle}.
  Let $d$ be the distance between each muscle, the weight of the edge is $-\beta{d}$, and the edges are connected throughout $\bm{x}$.
  $\beta$ is a coefficient for aligning the averages of $W$ and $\beta{d}$, and is automatically determined from the weights of all edges obtained.
}%
{%
  人間において空間的に近い筋ほど神経的に接続が強くなるという考えを, 空間的接続として取り込む.
  筋配置等の事前知識はないため, 本来であれば回路的に遅延が少ない, つまり空間的に近いという指標をロボットにおいては持ち込むべきであるが, 現状ではそれは難しい.
  そのため, 筋配置の情報は直接には使わないが, 幾何モデルにおけるそれぞれの筋経路の中心を計算し, それぞれの筋がそれぞれの筋に対してどの程度遠いか, を表すマトリックスのみ用いる.
  \figref{figure:spatial-muscle}のように, 手足を広げた状態において, このときの空間的な距離を近似的に空間的接続の距離として用いることとする.
  つまり, それぞれの筋同士の距離を$d$として, 辺の重みは$-\beta{d}$となり, 全$\bm{x}$同士に対してこの辺を張る.
  $\beta$は\secref{subsec:functional-muscle}において得られた$W$と$\beta{d}$の平均を揃えるための係数であり, 得られた全辺の重みから自動的に決まる.
}%

\section{Experiments} \label{sec:experiments}
\subsection{Experimental Setup} \label{subsec:experimental-setup}
\begin{figure}[t]
  \centering
  \includegraphics[width=0.8\columnwidth]{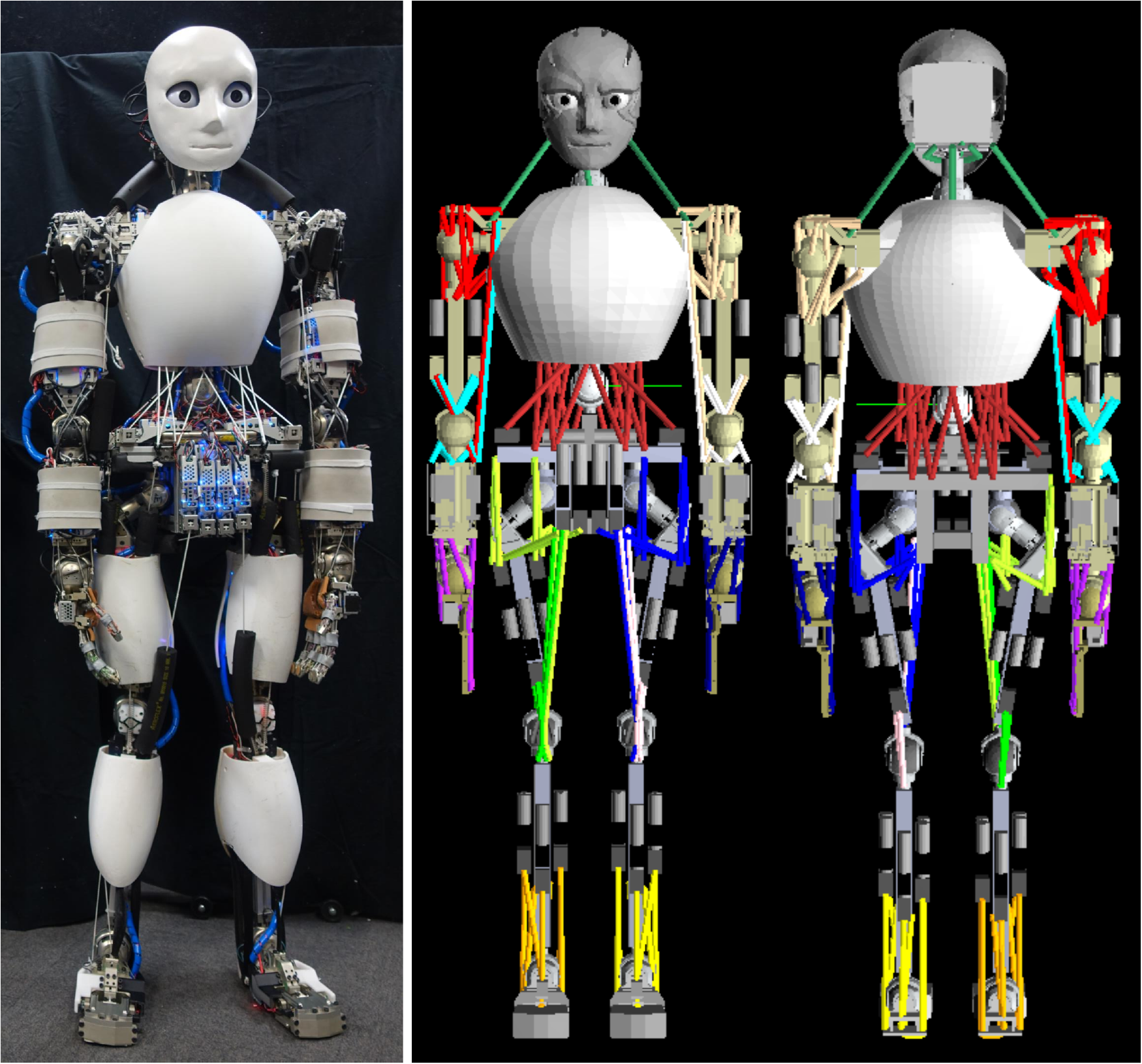}
  \vspace{-1.0ex}
  \caption{The musculoskeletal humanoid Musashi and its correct muscle grouping when using its geometric model (Geometric).}
  \label{figure:musashi-grouping}
  \vspace{-1.0ex}
\end{figure}

\begin{figure}[t]
  \centering
  \includegraphics[width=0.8\columnwidth]{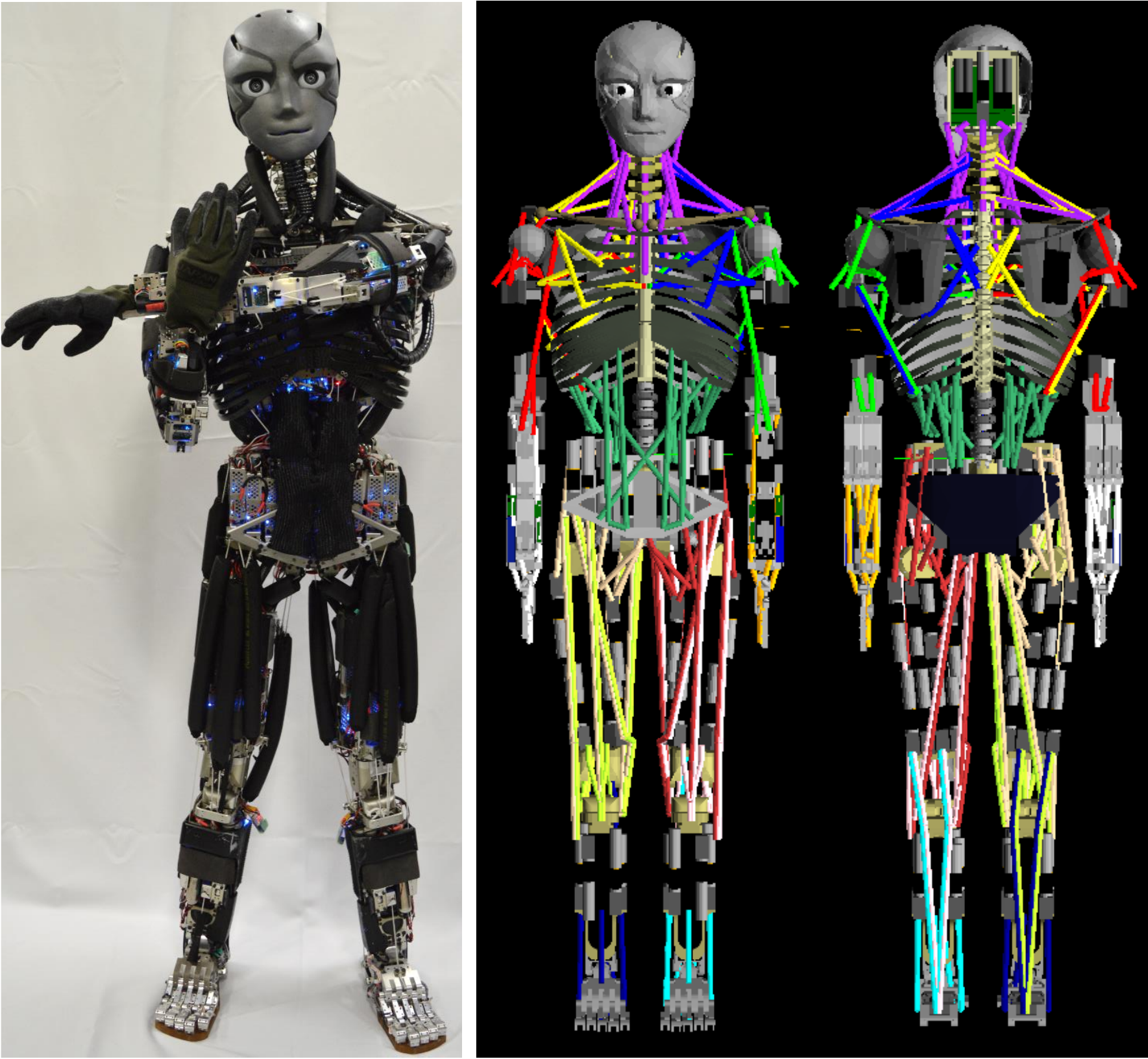}
  \vspace{-1.0ex}
  \caption{The musculoskeletal humanoid Kengoro and its correct muscle grouping when using its geometric model (Geometric).}
  \label{figure:kengoro-grouping}
  \vspace{-3.0ex}
\end{figure}

\switchlanguage%
{%
  The musculoskeletal humanoids used in this experiment are Musashi \cite{kawaharazuka2019musashi} and Kengoro \cite{asano2016kengoro}.
  The muscle arrangements are shown in \figref{figure:musashi-grouping} and \figref{figure:kengoro-grouping}.
  Musashi has 74 muscles and Kengoro has 116 muscles.
  Except for the foot and hands, Musashi has only four polyarticular muscles, while Kengoro has 26 polyarticular muscles, which is more similar to the human body.
  In \figref{figure:musashi-grouping} and \figref{figure:kengoro-grouping}, the muscles are grouped by color, which represents the groups of muscles used in the previous studies \cite{kawaharazuka2019longtime, kawaharazuka2020autoencoder}.
  In both cases, the number of groups is 14, and if a muscle has two colors, it means that the muscle is polyarticular and spans two groups.
  This grouping can be created by selecting a joint and then selecting all the muscles that contribute to the joint, which is the ground truth of this experiment.
  In contrast to this muscle grouping (Geometric), which is created based on a geometric model of joints and muscles, by using the proposed method, the robot performs the muscle grouping from its own random muscle movements without any geometric model of joints and muscles (Proposed).
  We will discuss the consistency between Geometric and Proposed muscle groupings, as well as the learning efficiency and accuracy before and after the grouping.

  Here, we calculate the consistency rate between Geometric and Proposed groupings, which is the ratio of whether or not a matched Proposed grouping is generated for each Geometric grouping, with $A_0$ representing the ratio of perfect consistency, $A_1$ representing the ratio that allows one different case, and $A_2$ representing the ratio that allows two different cases ($0 \leq A_{\{0, 1, 2\}} \leq 100$).
  For muscles that span two groups, it is acceptable to belong to either group.
}%
{%
  本実験で用いる筋骨格ヒューマノイドはMusashi \cite{kawaharazuka2019musashi}とKengoro\cite{asano2016kengoro}である.
  その筋配置を\figref{figure:musashi-grouping}, \figref{figure:kengoro-grouping}に示す.
  Musashiは74本, Kengoroは116本の筋を持つ.
  足(toe)と手(hand)を除き, Musashiは多関節筋が4本と少ないのに対して, Kengoroは26本と, より人体に似た筋配置が成されている.
  また, \figref{figure:musashi-grouping}, \figref{figure:kengoro-grouping}において筋が色分けされているが, これらは, これまで\cite{kawaharazuka2019longtime, kawaharazuka2020autoencoder}等で使われてきたmusashi, kengoroの筋のグルーピングを表す.
  どちらもグルーピング数は14であり, 一つの筋に2つの色分けが成されている場合は, どちらのグループにもまたがる多関節筋であるという意味である.
  これは, 関節を選び, その関節に寄与する筋を全て選択していくという方法により作成可能であり, 本実験のGround Truthとする.
  この事前に関節と筋配置に関する幾何モデルが得られる状態で作成された筋グルーピング(Geometric)に対して, 本実験では, 関節と筋配置の幾何モデルを前提としない状態で, ロボットが自身のランダムな筋の動きからこのグルーピングを行っていく(Proposed).
  GeometricとProposedにより得られた筋グルーピングの一致率, また, 分割前と後における学習効率・正確性の観点から議論を進める.

  ここで, GeometricとProposedによるグルーピングの一致率を算出する場合があるが, これは幾何モデルから得られたグルーピング一つずつに対して, それと一致するグルーピングが生成されたかどうかの割合を表し, 完全に一致する割合を$A_{0}$, 1つ異なる場合を許した割合を$A_{1}$, 2つ異なる場合を許した割合を$A_{2}$で表す$(0 \leq A_{\{0, 1, 2\}} \leq 100)$.
  2つのグルーピングにまたがる筋については, どちらのグループに属しても良いこととする.
}%

\begin{figure}[t]
  \centering
  \includegraphics[width=0.75\columnwidth]{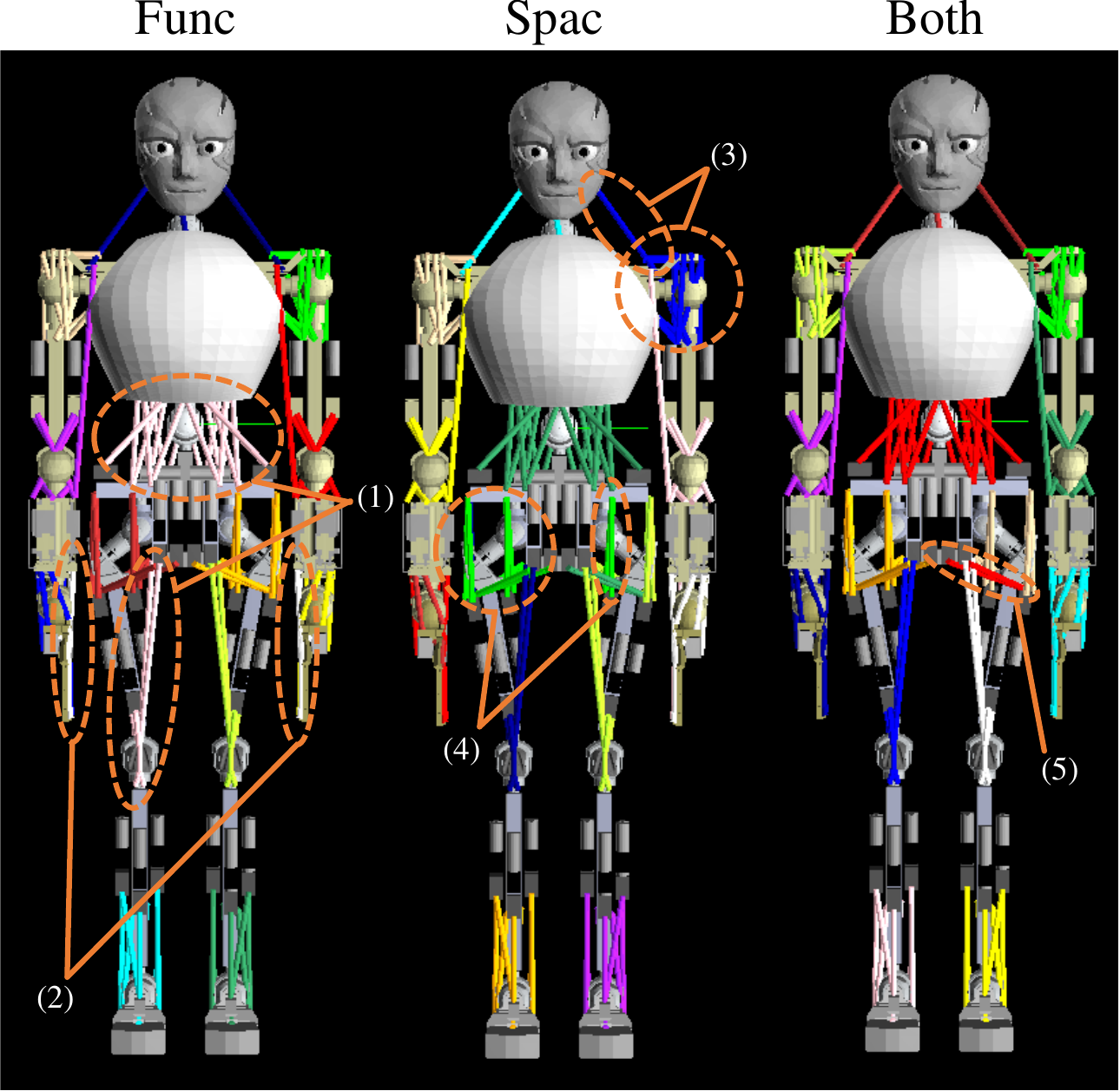}
  \vspace{-1.0ex}
  \caption{Examples of muscle grouping when conducting the proposed grouping method of Func, Spac, or Both for Musashi (Proposed).}
  \label{figure:musashi-sim}
  \vspace{-1.0ex}
\end{figure}

\begin{figure}[t]
  \centering
  \includegraphics[width=0.9\columnwidth]{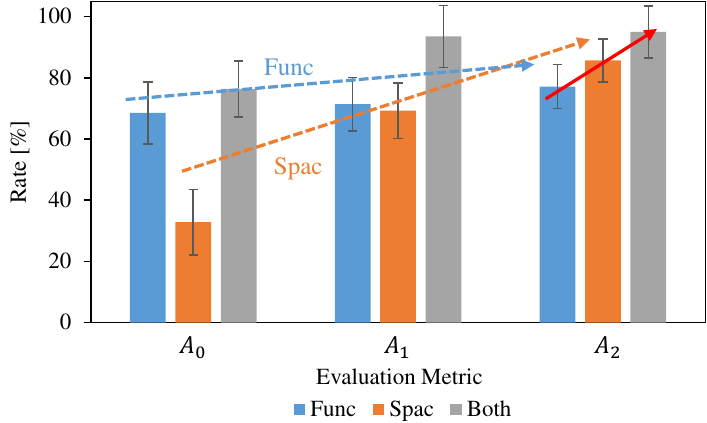}
  \vspace{-1.0ex}
  \caption{Evaluation metric of $A_0$, $A_1$, and $A_2$ when conducting the proposed grouping method of Func, Spac, or Both 10 times for Musashi.}
  \label{figure:musashi-eval}
  \vspace{-3.0ex}
\end{figure}

\begin{figure}[t]
  \centering
  \includegraphics[width=0.9\columnwidth]{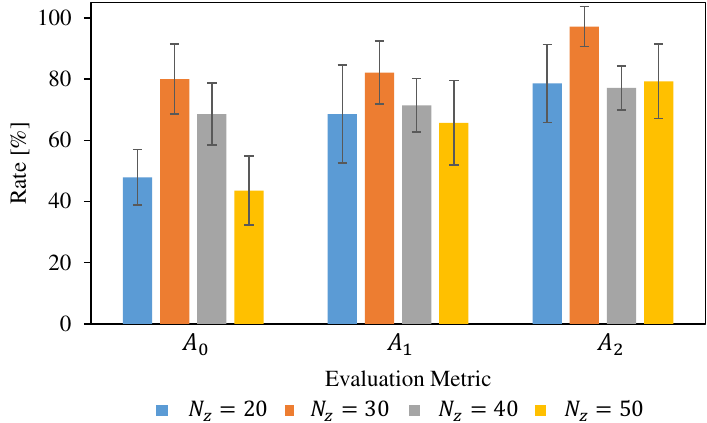}
  \vspace{-1.0ex}
  \caption{Comparison of $A_0$, $A_1$, and $A_2$ when changing $N_z$ to 20, 30, 40, or 50 and conducting the muscle grouping method of Func for Musashi.}
  \label{figure:musashi-z}
  \vspace{-1.0ex}
\end{figure}

\begin{figure}[t]
  \centering
  \includegraphics[width=1.0\columnwidth]{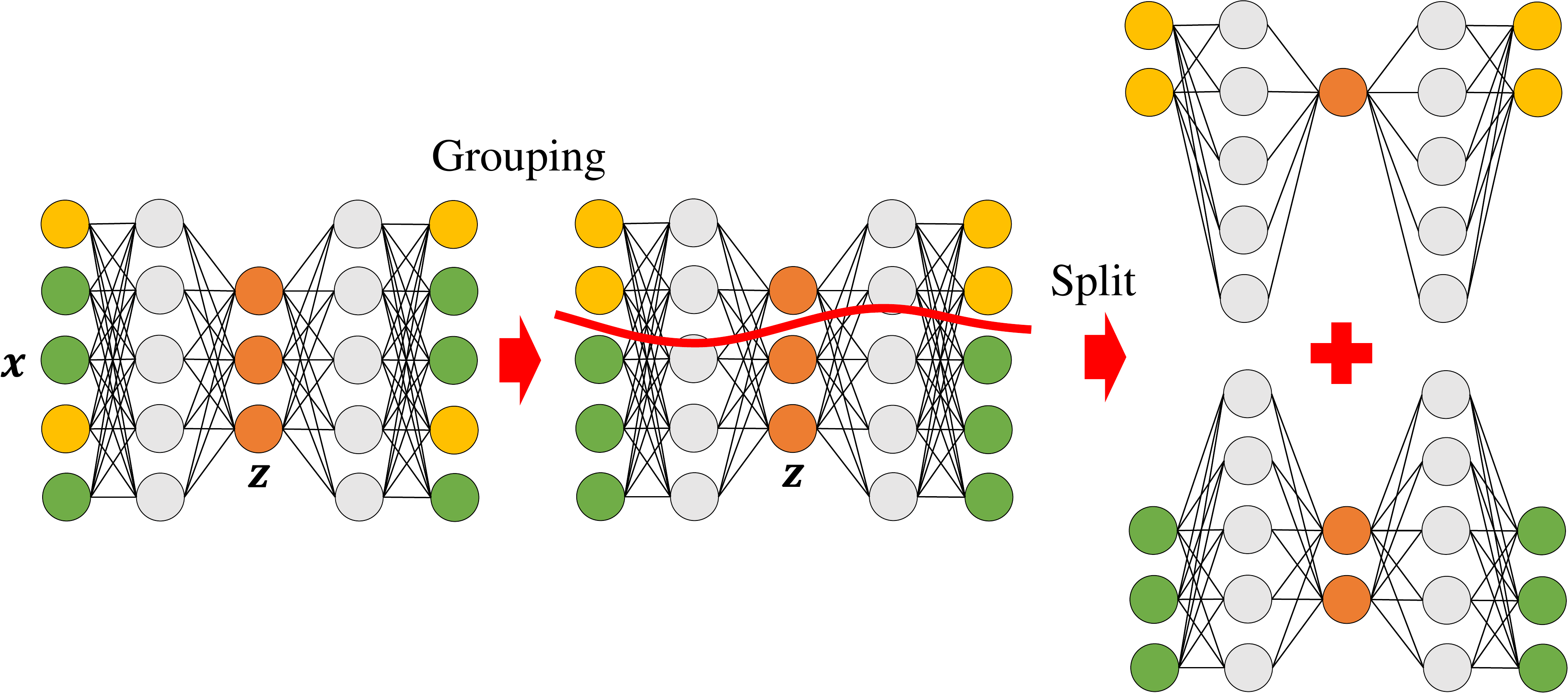}
  \vspace{-3.0ex}
  \caption{The split of AutoEncoder for the investigation of loss transition when training it before and after muscle grouping.}
  \label{figure:loss-divide}
  \vspace{-3.0ex}
\end{figure}

\begin{figure}[t]
  \centering
  \includegraphics[width=0.9\columnwidth]{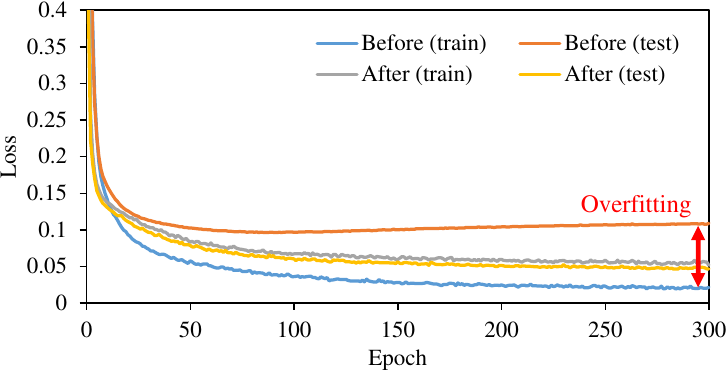}
  \vspace{-1.0ex}
  \caption{Transition of train/test loss when training AutoEncoder before and after muscle grouping for Musashi.}
  \label{figure:musashi-loss}
  \vspace{-3.0ex}
\end{figure}

\subsection{Evaluation Using Simulation} \label{subsec:simulation-evaluation}
\begin{figure}[t]
  \centering
  \includegraphics[width=0.75\columnwidth]{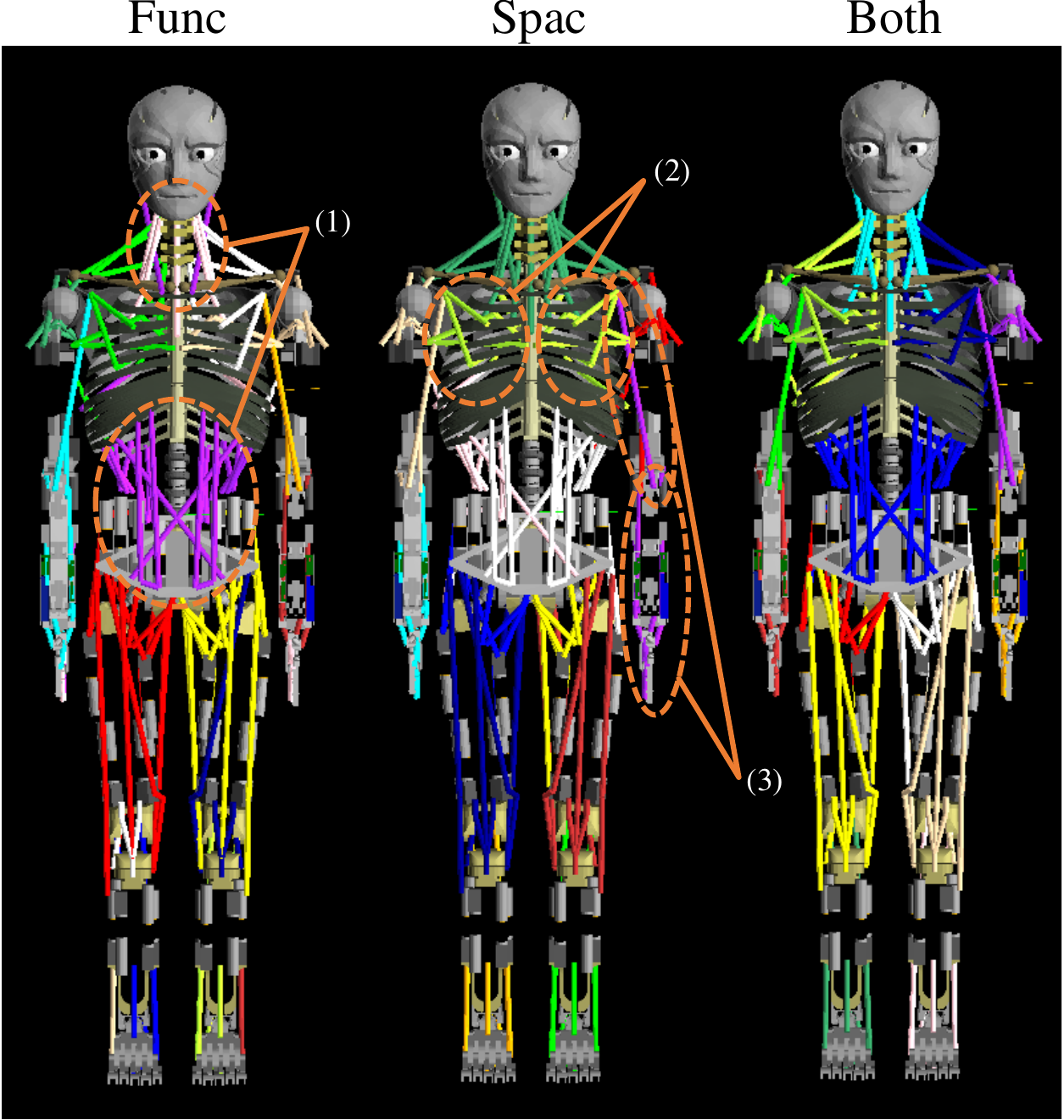}
  \vspace{-1.0ex}
  \caption{Examples of muscle grouping when conducting the proposed grouping method of Func, Spac, or Both for Kengoro (Proposed).}
  \label{figure:kengoro-sim}
  \vspace{-1.0ex}
\end{figure}

\begin{figure}[t]
  \centering
  \includegraphics[width=0.9\columnwidth]{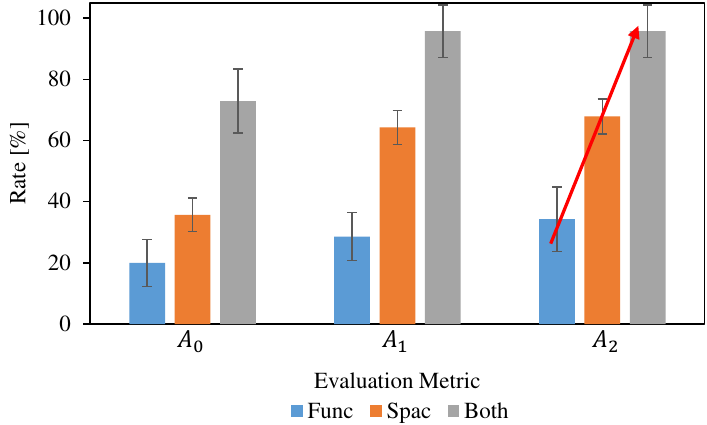}
  \vspace{-1.0ex}
  \caption{Evaluation metric of $A_0$, $A_1$, and $A_2$ when conducting the proposed grouping method of Func, Spac, or Both 10 times for Kengoro.}
  \label{figure:kengoro-eval}
  \vspace{-3.0ex}
\end{figure}

\switchlanguage%
{%
  The simulation here refers to the use of a human-made geometric model of a muscle path linearly connecting the start, relay, and end points of the muscle.
  The muscle grouping is based on only two pieces of data: the 100,000 muscle lengths $\bm{l}$ of all muscles in random postures within the joint angle range, and information of spatial connections described in \secref{subsec:spatial-muscle}.
  Here, we add noise with an average of 0 and a standard deviation of 100 mm for the distance $d$ between the muscles in this study, because obtaining the information of spatial connections from the geometric model is unrealistic.
  The data of $\bm{l}$ are converted to functional connections by training an AutoEncoder as described in \secref{subsec:functional-muscle}.
  In this study, the AutoEncoder has five layers, the numbers of units are $M$, $300$, $N_{z}$, $300$, and $M$ in order (where $M$ is the number of muscles), the activation function is hyperbolic tangent, and the batch normalization \cite{ioffe2015batchnorm} is applied to each layer except the final layer.
  For $N_{z}$, we try several values and compare them.
  Also, we divide the data into two parts: 80\% for training and 20\% for testing, the number of batches is set to 100, the number of epochs is set to 300, the update rule is set to Adam \cite{kingma2015adam}, and the model with the lowest test value is used.
  We compare the grouping performance of Musashi and Kengoro in the case of using only functional connection (Func), using only spatial connection (Spac), and both connections (Both).

  Examples of the muscle grouping in Musashi are shown in \figref{figure:musashi-sim}.
  Here, with $N_{z}=40$, we show the overall figure and details of the grouping of Func, Spac, and Both, in order.
  As for the color of each group, the same group may have different colors due to different initial seeds of grouping.
  For Func, Spac and Both, the grouping is roughly the same as in \figref{figure:musashi-grouping}.
  In the case of Func, groups from relatively distant locations shown in (1) and (2) of \figref{figure:musashi-sim}, such as the knee and hip, and the right and left hand, sometimes belong to the same group.
  On the other hand, in the case of Spac, the wrong groupings are often found as in the case of (3) and (4) of \figref{figure:musashi-sim}, where two spatially close but functionally unrelated groups are combined in the same group, such as the neck and shoulder or the right and left hip.
  In the case of Both, although the same grouping as Geometric grouping is generated with high probability, some muscles are sometimes scattered to other groups as shown in (5) of \figref{figure:musashi-sim}.
  The mean and variance of $A_{0}$, $A_{1}$, and $A_{2}$ after 10 grouping trials are shown in \figref{figure:musashi-eval}.
  From the results, Both is most consistent with Geometric for all of $A_0$, $A_1$, and $A_2$.
  For Func, the consistency rate does not change among $A_0$, $A_1$ and $A_2$, while for Spac, the rate increases significantly from $A_0$ to $A_2$.
  The reason for this is that in the case of Func, it is meaningless to allow a few errors because Func has multiple muscles spanning two distant groups, while in the case of Spac, only one or two spatially close wrong muscles often belong to the same group.

  Next, the mean and variance of $A_{0}$, $A_{1}$, and $A_{2}$ for Func with $N_{z}$ being changed to 20, 30, 40, and 50 are shown in \figref{figure:musashi-z}.
  For all metrics, $N_z=30$ is the best, and higher or lower value reduces the consistency rate.
  The number of joints relating with all muscles in Musashi is 46, and $N_z=30$ is much smaller than that.
  For Both, when $N_{z}=\{20, 30, 40, 50\}$, $A_{2}=\{97, 97, 95, 87\}$\%, and there is no significant change compared to Func when $N_{z}$ is small enough.

  Before and after this grouping, we investigated how the loss transition changes when training the AutoEncoder.
  In this experiment, the number of data is reduced to 1000, making training difficult and prone to overfitting.
  Regarding after the muscle grouping, we use the result of Both in \figref{figure:musashi-sim}, split $\bm{l}$ and $\bm{z}$ as shown in \figref{figure:loss-divide}, and train AutoEncoder.
  Here, the total number of weights is the same before and after the muscle grouping.
  The loss after the muscle grouping is an average of the losses from each group, weighted by the size of $\bm{l}$ in each group.
  The loss transitions are shown in \figref{figure:musashi-loss}.
  Before the muscle grouping, the train loss is significantly lower than the test loss, thus overfitted, but after the grouping, the train loss is not overfitted.
  The overfitting is considered to be reduced by the disappearance of unrelated extra variables.

  An example of the muscle grouping in Kengoro is shown in \figref{figure:kengoro-sim}.
  Here, we set $N_{z}=50$.
  Kengoro has a more complex body structure than Musashi, but as in the case of Musashi, two distant unrelated groups often merge in Func, while in Spac, two functionally unrelated groups that are close to each other often merge or only one unrelated muscle is grouped into a spatially close group.
  The mean and variance of $A_{0}$, $A_{1}$, and $A_{2}$ after 10 grouping trials are shown in \figref{figure:kengoro-eval}.
  The trend is similar to that of Musashi, but the consistency rates of Func and Spac are lower than those of Musashi, which shows that Kengoro has more complex body structures.
  The same performance as Musashi is achieved by using both functional and spatial connections.
}%
{%
  ここでいうシミュレーションとは, 人間が作った, 筋経路を筋の始点・中継点・終点を直線で結んだ幾何モデルを用いることを指す.
  幾何モデルを関節角度範囲内でランダムな姿勢にしたときの全筋の筋長$\bm{l}$を並べた100000のデータと, \secref{subsec:spatial-muscle}における空間的接続の情報の2つのみを元に筋分割を行う.
  ここで, 筋のまとまりがはっきりしている場合には空間的接続の情報は強すぎる条件となってしまうため, 本研究では筋同士の距離$d$に対して, 平均0, 分散100 mmのノイズを加えている.
  $\bm{l}$のデータは, \secref{subsec:functional-muscle}に述べたようにAutoEncoderを学習することで機能的接続に変換する.
  本研究では, このAutoEncoderは5層とし, ユニット数は順に$M, 300, N_{z}, 300, M$ (ここで$M$は筋の数とする), 活性化関数はtanh, 最終層以外のそれぞれの層についてbatch normalization \cite{ioffe2015batchnorm}を適用している.
  $N_{z}$については, いくつかの値を試し比較を行う.
  また, データは80\%を訓練用, 20\%をテスト用に分け, バッチ数を100, エポック数を300, 更新則をAdam \cite{kingma2015adam}として学習を行い, 最もテストの値が小さいモデルを使用する.
  機能的接続のみを用いた場合(Func), 空間的接続のみを用いた場合(Spac), 両者を用いた場合(Both)のグルーピング性能について, Musashi, Kengoroの両者それぞれについて比較を行う.

  Musashiにおける分割の例を\figref{figure:musashi-sim}に示す.
  ここでは$N_{z}=40$としており, 順にFunc, Spac, Bothにおける分割の全体像とその詳細を示す.
  それぞれのグルーピングの色については, 初期値が異なるため同じグループでも異なる色となることがある.
  Func, Spac, Bothそれぞれについて, 概ね\figref{figure:musashi-grouping}と同じようなグルーピングが成されていた.
  Funcでは, 図に示した(1), (2)のように, 膝と腰, 右手と左手のような, 比較的遠い別であるべき場所のグループ同士が合体して同じグループに属してしまうことがあった.
  これに対して, Spacについては, 図に示した(3), (4)のように首と肩や右と左の股関節のような, 空間的には近いが機能的には干渉が無く別のグループになるべき場所において, 一本だけ間違ったグルーピングが成されていることが多い.
  Bothについては高い確率でGeometricと同じグルーピングが作成されたが, (5)のように一部別のグループに筋が飛んでしまうことがあった.
  このグルーピングを10回行った際における, $A_{0}, A_{1}, A_{2}$の平均と分散を\figref{figure:musashi-eval}に示す.
  \figref{figure:musashi-eval}から分かるように, $A_0, A_1, A_2$の全てにおいて機能的接続と空間的接続を用いるBothによるグルーピングが最もGeometricと一致していた.
  Funcについては$A_0, A_1, A_2$について一致率にはほとんど変化がないのに対して, Spacについては$A_0$から$A_2$になるに従って大きく一致率が向上していた.
  これは, Funcは離れたグループにまたがって属する複数の筋を持つため, 間違いを数本許容しても意味がないのに対して, Spacでは一本や二本だけ間違った筋が近いグルーピングに属してしまうことが多かったためだと考えられる.

  次に, Musashiにおいて, $N_{z}$を20, 30, 40, 50に変化させたときのFuncに関する$A_{0}, A_{1}, A_{2}$の平均と分散を\figref{figure:musashi-z}に示す.
  全てのmetricにおいて$N_z=30$が最も良く, それ以上高くても低くても, 一致率が減少することが分かった.
  Musashiの全筋に関わる関節自由度数は46であり, その値よりもかなり小さな値であった.
  また, Bothについては, $N_{z}=\{20, 30, 40, 50\}$のとき, $A_{2}=\{97, 97, 95, 87\}$\%と, $N_{z}$が十分小さな場合はFuncと比べて大きな変化は見られなかった.

  この分割をする前と後について, AutoEncoderを学習させた際のlossの変化はどの程度変わるかについて検証する.
  この際はデータ数は1000まで落とし, 学習が難しく, 過学習しやすい状態となっている.
  分割後のグルーピングについては, \figref{figure:musashi-sim}のBothの結果について行い, 分割されたそれぞれのグループの$\bm{l}$と$\bm{z}$を用いて同様にAutoEncoderを学習させてみる.
  このとき, 分割前と分割後で重みの総数は変わらない.
  分割後のlossについては, それぞれのグループから得られたlossを$\bm{l}$の大きさで重み付け平均したものである.
  lossの遷移を\figref{figure:musashi-loss}に示す.
  分割前はtrain lossがtest lossを大きく下回り過学習しているのに対して, 分割後は過学習していない.
  関係のない余計な変数が消えることで, 過学習しにくくなったと考えられる.

  Kengoroにおける分割の例を\figref{figure:kengoro-sim}に示す.
  ここでは$N_{z}=50$としている.
  Kengoroの方がMusashiよりも身体構造は複雑であるが, Musashiのときと同様に, Funcでは比較的離れた2つのグループが合体することが多く, Spacでは近い2つのグループが合体, または一本だけ近い異なる方へグルーピングされてしまうことが多かった.
  このグルーピングを10回行った際における, $A_{0}, A_{1}, A_{2}$の平均と分散を\figref{figure:kengoro-eval}に示す.
  傾向はMusashiと似ているが, FuncとSpacについてはMusashiよりも一致率が低く, より複雑で難しい身体構造であることがわかる.
  同時に, Bothでは機能的接続と空間的接続を併用することで, Musashiの場合と同等の性能を叩き出している.
}%

\begin{figure}[t]
  \centering
  \includegraphics[width=1.0\columnwidth]{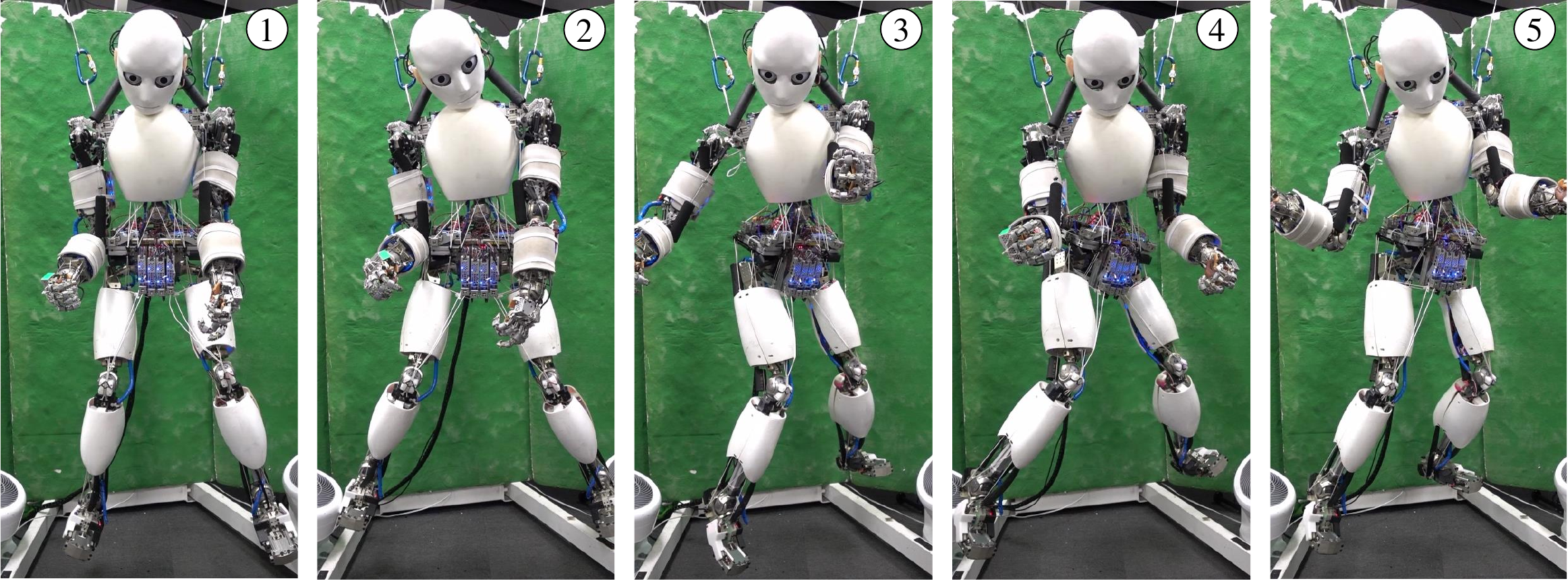}
  \vspace{-3.0ex}
  \caption{The experiment of collecting muscle length data from random movements of the actual robot Musashi.}
  \label{figure:actual-experiment}
  \vspace{-1.0ex}
\end{figure}

\begin{figure}[t]
  \centering
  \includegraphics[width=0.9\columnwidth]{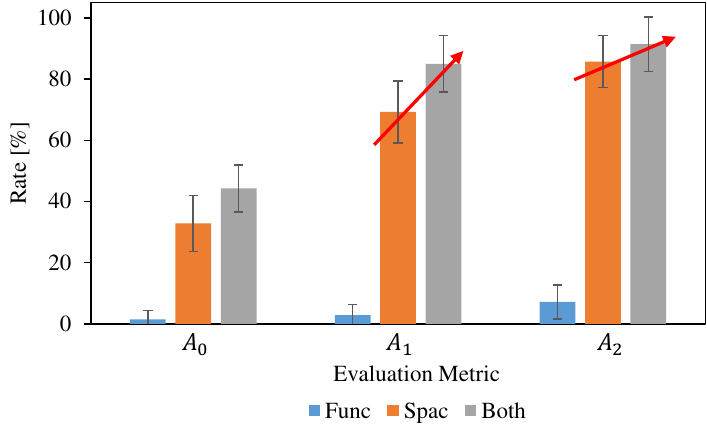}
  \vspace{-1.0ex}
  \caption{Evaluation metric of $A_0$, $A_1$, and $A_2$ when conducting the proposed grouping method of Func, Spac, or Both 10 times for the actual robot Musashi.}
  \label{figure:musashi-actual}
  \vspace{-3.0ex}
\end{figure}

\subsection{Evaluation Using the Actual Robot Musashi} \label{subsec:actual-evaluation}
\switchlanguage%
{%
  This experiment is conducted using the actual robot of the musculoskeletal humanoid Musashi.
  The same method as with \secref{subsec:simulation-evaluation} is used for spatial connection, but the data sequence of $\bm{l}$ is obtained from actual sensor data of random movements.
  Since no joint or muscle arrangement information is used, a range of muscle length is determined for each muscle and a random target muscle length is sent to the robot (\figref{figure:actual-experiment}).
  Here, the maximum and minimum muscle tensions are controlled for each muscle in the same way with \cite{kawaharazuka2020thermo}, in order to suppress the phenomenon of the antagonistic muscles pulling or loosening each other.
  By running the data collection at 2Hz for about 25 minutes, about 3000 muscle lengths are obtained and a functional connection is obtained in the same way as \secref{subsec:simulation-evaluation}.
  Since the number of data is small, we set the number of batches to 50, the number of epochs to 3000, and $N_z=40$.
  The mean and variance of $A_{0}$, $A_{1}$, and $A_{2}$ after 10 grouping trials by Func, Spac, and Both are shown in \figref{figure:musashi-actual}.
  The grouping with Func is difficult and almost always unsuccessful.
  On the other hand, Both improves the consistency rate by around 10-20\% compared to Spac by using a functional connection.
  Although the data collection with the actual robot is greatly inferior to that with the geometric model, because the number of data is small due to the difficulty of data acquisition on the actual robot and the large noise, we can see that the accuracy of muscle grouping can be increased by using the collected data together with the spatial connection.
}%
{%
  筋骨格ヒューマノイドMusashiの実機を使って実験を行う.
  空間的接続については\secref{subsec:simulation-evaluation}と同じものを用いるが, $\bm{l}$のデータ列を実機のランダムなデータから取得する.
  関節や筋配置の情報を一切用いないため, それぞれの筋について長さの範囲を決め, ランダムな筋長を送る(\figref{figure:actual-experiment}).
  このとき, 拮抗筋同士が引っ張りあい強い筋張力を発揮したり互いに緩む現象を抑制するため, \cite{kawaharazuka2020thermo}と同様の方法で筋張力の最大値と最小値を筋ごと設定している.
  2Hzで約25分間動かして, 約3000のデータを取得し, \secref{subsec:simulation-evaluation}と同様に機能的接続を得る.
  ここで, 学習の際はデータが少ないためbatch数を50, epoch数を3000とし, $N_z=40$とする.
  このときのFunc, Spac, Bothによる分割をそれぞれ10回行った際における$A_{0}, A_{1}, A_{2}$の平均と分散を\figref{figure:musashi-actual}に示す.
  Funcによる分割は難しく, ほとんど成功しなかった.
  これに対して, Bothでは, 機能的接続を用いることでSpacよりも10-20\%前後も一致率が上がった.
  これは, 実機ではデータ取得が難しいためデータ数が少なくノイズも大きいため, 幾何モデルと比べると大きく劣るが, 空間的接続と併用することで, 精度を上げられるということだと考えられる.
}%

\section{Discussion} \label{sec:discussion}
\switchlanguage%
{%
  From this experiment, it is found that the combination of functional and spatial connections enables the accurate grouping of muscles.
  The functional connection alone causes spatially distant muscles to belong to the same group, while the spatial connection alone causes spatially close but functionally different muscles to be grouped together.
  By combining these two, the simulation results for Musashi and Kengoro show that $A_0$ is about 80\% and $A_2$ is about 95\%.
  Since it is difficult to obtain a large amount of data on the actual robot, functional connections alone are not sufficient for grouping, but it can be improved by combining them with spatial connections.
  As for the AutoEncoder in obtaining functional connections, the dimension of $\bm{z}$ has a proper value, and it is smaller than the number of joints that should be correct as $N_{z}$.

  The muscle grouping method is chosen as the target of this method since it is easy to evaluate because of the existence of the ground truth, which can be obtained from the arrangement of joints and their related muscles if a geometric model is known.
  In future works, we will be able to construct a robot that can gradually understand the relationship between the muscles and the joints from random movements, and will be able to construct a robot that has organized muscles without giving any information on muscle or joint arrangement.
  In this study, the number of groups is limited to 14 for the purpose of evaluation.
  When the number of groups is reduced, the groups that are closely related to each other are merged, and when the number of groups is increased, the muscles are further divided by each degrees of freedom of the joints.
  However, in practice, it is necessary to try several numbers of groups, to create controllers and recognizers with it, and to decide the number of groups based on the trade-offs of accuracy and robustness.
  This is task-dependent and therefore difficult to evaluate, but we need to work on it in the future.

  Our method is applicable to the case where the functions of sensors and actuators are clearly divided, and so the performance is likely to be worse when the functions are very closely connected.
  The effects on the controller and state estimator when some groupings are wrong, and the grouping method considering the sensors and actuators across multiple groups are major issues to be addressed in the future.
  In addition, a formal verification of our method is also important, and we would like to conduct it in a way that is consistent with the actual robot system containing many noises.
  This method can be used not only for musculoskeletal structures but also for contact sensors, inertial sensors, temperature sensors, etc. distributed throughout the whole body.
  Also, it is possible to use our method not only for one-dimensional sensors like muscle lengths but also for multi-dimensional sensors like three-axis tactile sensors, considering each axis as a single sensor.
  We are convinced that our method will be useful when contact sensors that have been implemented only in the fingers or special sensors that have been implemented only in the arms are implemented in the whole body.
  Also, although only the static relationship is handled in this study, when the dynamic relationship is strong, it is necessary to add differential information of the sensors or to make the AutoEncoder recurrent by using LSTM.

  This method automatically organizes the muscles and automates the process of constructing neural networks such as \cite{kawaharazuka2020autoencoder}.
  Once the robot is assembled, even if it is flexible and difficult to modelize, it acquires its own body image by moving at random and becomes able to achieve tasks gradually.
}%
{%
  本実験から, 機能的接続と空間的接続を併用することで, 精度高く筋群をグルーピングすることが可能であることがわかった.
  機能的接続だけでは空間的に遠い筋同士が同じグループに属してしまい, 空間的接続だけでは空間的には違いが機能的には異なる筋同士を同じグループにしてしまう.
  この2つを合わせることで, SimulationではMusashi, Kengoroについて$A_0$は80\%前後, $A_2$は95\%前後を達成することができた.
  また, 実機では大量のデータを取得することが難しいため, 機能的接続だけでは上手くグルーピングすることが出来なかったが, 空間的接続と合わせることで一致率を向上させることが可能であった.
  機能的接続を獲得する際のAutoEncoderについては, $\bm{z}$の次元に適切な値が存在し, これは本来のあるべき次元である関節自由度数よりも小さい.

  筋分割は, 幾何モデルが分かれば関節とそれに関わる筋の配置から行えるため, ground truthがあり評価がしやすいため本手法の適用先に選んだ.
  今後, 筋配置や関節配置等を与えなくても, ランダムな動きから徐々にそれらの関係が分かり, 筋が組織化されていくロボットが構成できると考える.
  また, 今回は評価のためにグルーピング数を14と限定した.
  分割数をより小さくすると, 関係の深いグルーピング同士が合体したようなグループが生成され, 大きくするとさらに関節の持つ自由度数ごとに筋が分割されていくことになる.
  しかし, 実際には分割数を限定せず, 数種類を試し, 分割した状態で制御器や認識器を作成し, その正確性やロバスト性のトレードオフから分割数を決定していく必要がある.
  これはタスクに依存するため評価は難しいが, 今後取り組んでいく必要がある.

  本手法は, 機能が明確にわかれやすいようなものに対して適用できる手法であり, それぞれが非常に蜜に繋がっている場合には性能が落ちる可能性が高い.
  しかし, 機能がわかれたものについては良い性能を示し, 筋構造だけでなく, 全身に分布した接触センサや慣性センサ, 温度センサ等に使うことができると考える.
  今後, 指だけに実装された接触センサや, 腕だけに実装された特殊なセンサ群が, 全身となったとき, その助けとなると確信する.
  また, 今回は静的な関係だけを見たが, 動的な関係が強い場合には, 速度情報を入れてあげたり, AutoEncoderをLSTM等を用いてリカレントにする必要がある.

  本手法により, 自動的に筋が組織化され, その中で\cite{kawaharazuka2020autoencoder}のような学習器を構成していくという工程が自動化される.
  例え柔軟でモデル化が難しいロボットでも, 組み上がったら, ランダムに動きながら自動で自身の身体図式を獲得し, 徐々に動けるようになっていく, そんな手法の一助になれば幸いである.
}%

\section{CONCLUSION} \label{sec:conclusion}
\switchlanguage%
{%
  In this study, we proposed an automatic grouping method of the redundant sensors and actuators based on their functional and spatial connections.
  By acquiring the functional connections using AutoEncoder and constructing a graph structure with the spatial connections, the randomized selection algorithm allows us to divide the sensors and actuators into the groups with few connections.
  As a concrete example, we considered the muscle grouping of the musculoskeletal humanoid, and the functional relationship from the antagonism of the muscles and the spatial connections of the neural connections are embedded into a relational graph.
  Without any information on the joint and muscle arrangement, this method enables us to group the muscles into regions such as the forearm, upper arm, neck, and waist with awareness of the antagonistic relationship.
  Compared with the neural network before the grouping, after the grouping, the proposed method is able to obtain an interpretable and hard to overfit structure while maintaining some degree of accuracy.
  In future works, we will continue to apply this method to the task-based environment, in order to improve the efficiency of learning, online learning, and network interpretation.
}%
{%
  本研究では, 冗長なセンサ・アクチュエータを全身に持つロボットについて, その機能的接続と空間的な接続から, それらを分割する手法について提案した.
  機能的接続をAutoEncoderを用いて獲得し, 空間的接続と合わせてグラフ構造を作成することで, 乱択アルゴリズムにより, 関わりの少ないグルーピングごとに分割することができる.
  具体例として筋骨格ヒューマノイドの筋分割を挙げ, 筋の拮抗関係からくる機能的な関係と, 神経接続からくる空間的な接続をグラフに落とし込むことを行った.
  関節や筋配置の情報がない状態で, 本手法によって拮抗関係を意識しつつ前腕や上腕, 首や腰等の部位ごとに, 筋を分割することが可能となった.
  分割する前に比べ, 分割後は正確性をある程度保ったまま, 解釈性の向上した, 過学習しにくい構造を獲得できることを示した.
  今後は, 本手法をタスクをベースに適用していき, 学習の効率化・オンライン学習・ネットワークの解釈性向上を目指していきたい.
}%

\section*{Acknowledgement}
This research was supported by JST ACT-X Grant Number JPMJAX20A5 and JSPS KAKENHI Grant Number JP19J21672.
The authors would like to thank Yuka Moriya for proofreading this manuscript.

{
  \bibliographystyle{IEEEtran}
  \bibliography{main}

\begin{thebibliography}{10}
\providecommand{\url}[1]{#1}
\csname url@rmstyle\endcsname
\providecommand{\newblock}{\relax}
\providecommand{\bibinfo}[2]{#2}
\providecommand\BIBentrySTDinterwordspacing{\spaceskip=0pt\relax}
\providecommand\BIBentryALTinterwordstretchfactor{4}
\providecommand\BIBentryALTinterwordspacing{\spaceskip=\fontdimen2\font plus
\BIBentryALTinterwordstretchfactor\fontdimen3\font minus
  \fontdimen4\font\relax}
\providecommand\BIBforeignlanguage[2]{{%
\expandafter\ifx\csname l@#1\endcsname\relax
\typeout{** WARNING: IEEEtran.bst: No hyphenation pattern has been}%
\typeout{** loaded for the language `#1'. Using the pattern for}%
\typeout{** the default language instead.}%
\else
\language=\csname l@#1\endcsname
\fi
#2}}

\bibitem{duan2016benchmarking}
Y.~Duan, X.~Chen, R.~Houthooft, J.~Schulman, and P.~Abbeel, ``{Benchmarking
  deep reinforcement learning for continuous control},'' in \emph{Proceedings
  of the 33rd International Conference on Machine Learning}, 2016, pp.
  1329--1338.

\bibitem{kawaharazuka2020autoencoder}
K.~Kawaharazuka, K.~Tsuzuki, M.~Onitsuka, Y.~Asano, K.~Okada, K.~Kawasaki, and
  M.~Inaba, ``{Musculoskeletal AutoEncoder: A Unified Online Acquisition Method
  of Intersensory Networks for State Estimation, Control, and Simulation of
  Musculoskeletal Humanoids},'' \emph{IEEE Robotics and Automation Letters},
  vol.~5, no.~2, pp. 2411--2418, 2020.

\bibitem{mittendorfer2011tactile}
P.~Mittendorfer and G.~Cheng, ``{Humanoid Multimodal Tactile-Sensing
  Modules},'' \emph{IEEE Transactions on Robotics}, vol.~27, no.~3, pp.
  401--410, 2011.

\bibitem{asano2016kengoro}
Y.~Asano, T.~Kozuki, S.~Ookubo, M.~Kawamura, S.~Nakashima, T.~Katayama,
  Y.~Iori, H.~Toshinori, K.~Kawaharazuka, S.~Makino, Y.~Kakiuchi, K.~Okada, and
  M.~Inaba, ``{Human Mimetic Musculoskeletal Humanoid Kengoro toward Real World
  Physically Interactive Actions},'' in \emph{Proceedings of the 2016 IEEE-RAS
  International Conference on Humanoid Robots}, 2016, pp. 876--883.

\bibitem{kawaharazuka2019musashi}
K.~Kawaharazuka, S.~Makino, K.~Tsuzuki, M.~Onitsuka, Y.~Nagamatsu, K.~Shinjo,
  T.~Makabe, Y.~Asano, K.~Okada, K.~Kawasaki, and M.~Inaba, ``{Component
  Modularized Design of Musculoskeletal Humanoid Platform Musashi to
  Investigate Learning Control Systems},'' in \emph{Proceedings of the 2019
  IEEE/RSJ International Conference on Intelligent Robots and Systems}, 2019,
  pp. 7294--7301.

\bibitem{kawamura2016jointspace}
M.~Kawamura, S.~Ookubo, Y.~Asano, T.~Kozuki, K.~Okada, and M.~Inaba, ``{A
  Joint-Space Controller Based on Redundant Muscle Tension for Multiple DOF
  Joints in Musculoskeletal Humanoids},'' in \emph{Proceedings of the 2016
  IEEE-RAS International Conference on Humanoid Robots}, 2016, pp. 814--819.

\bibitem{kawaharazuka2019longtime}
K.~Kawaharazuka, K.~Tsuzuki, S.~Makino, M.~Onitsuka, Y.~Asano, K.~Okada,
  K.~Kawasaki, and M.~Inaba, ``{Long-time Self-body Image Acquisition and its
  Application to the Control of Musculoskeletal Structures},'' \emph{IEEE
  Robotics and Automation Letters}, vol.~4, no.~3, pp. 2965--2972, 2019.

\bibitem{kawaharazuka2018estimator}
K.~Kawaharazuka, S.~Makino, M.~Kawamura, Y.~Asano, K.~Okada, and M.~Inaba, ``{A
  Method of Joint Angle Estimation Using Only Relative Changes in Muscle
  Lengths for Tendon-driven Humanoids with Complex Musculoskeletal
  Structures},'' in \emph{Proceedings of the 2018 IEEE-RAS International
  Conference on Humanoid Robots}, 2018, pp. 1128--1135.

\bibitem{jantsch2011scalable}
M.~J{\"a}ntsch, C.~Schmaler, S.~Wittmeier, K.~Dalamagkidis, and A.~Knoll, ``{A
  scalable joint-space controller for musculoskeletal robots with spherical
  joints},'' in \emph{Proceedings of the 2011 IEEE International Conference on
  Robotics and Biomimetics}, 2011, pp. 2211--2216.

\bibitem{kawaharazuka2020dynamics}
K.~Kawaharazuka, K.~Tsuzuki, M.~Onitsuka, Y.~Asano, K.~Okada, K.~Kawasaki, and
  M.~Inaba, ``{Object Recognition, Dynamic Contact Simulation, Detection, and
  Control of the Flexible Musculoskeletal Hand Using a Recurrent Neural Network
  With Parametric Bias},'' \emph{IEEE Robotics and Automation Letters}, vol.~5,
  no.~3, pp. 4580--4587, 2020.

\bibitem{niiyama2010emg}
R.~Niiyama, S.~Nishikawa, and Y.~Kuniyoshi, ``{Athlete Robot with applied human
  muscle activation patterns for bipedal running},'' in \emph{Proceedings of
  the 2010 IEEE-RAS International Conference on Humanoid Robots}, 2010, pp.
  498--503.

\bibitem{olsson2004grouping}
L.~Olsson, C.~L. Nehaniv, and D.~Polani, ``{Sensory channel grouping and
  structure from uninterpreted sensor data},'' in \emph{Proceedings of the 2004
  NASA/DoD Conference on Evolvable Hardware}, 2004, pp. 153--160.

\bibitem{hinton2006reducing}
G.~E. Hinton and R.~R. Salakhutdinov, ``{Reducing the Dimensionality of Data
  with Neural Networks},'' \emph{Science}, vol. 313, no. 5786, pp. 504--507,
  2006.

\bibitem{nagamochi2002mincut}
H.~Nagamochi and T.~Ibaraki, ``{Graph connectivity and its augmentation:
  applications of MA orderings},'' \emph{Discrete Applied Mathematics}, vol.
  123, no.~1, pp. 447--472, 2002.

\bibitem{kruskal1956spanning}
J.~B. Kruskal, ``{On the shortest spanning subtree of a graph and the traveling
  salesman problem},'' \emph{Proceedings of the American Mathematical society},
  vol.~7, no.~1, pp. 48--50, 1956.

\bibitem{ioffe2015batchnorm}
S.~Ioffe and C.~Szegedy, ``{Batch Normalization: Accelerating Deep Network
  Training by Reducing Internal Covariate Shift},'' in \emph{Proceedings of the
  32nd International Conference on Machine Learning}, 2015, pp. 448--456.

\bibitem{kingma2015adam}
D.~P. Kingma and J.~Ba, ``{Adam: A Method for Stochastic Optimization},'' in
  \emph{Proceedings of the 3rd International Conference on Learning
  Representations}, 2015, pp. 1--15.

\bibitem{kawaharazuka2020thermo}
K.~Kawaharazuka, N.~Hiraoka, K.~Tsuzuki, M.~Onitsuka, Y.~Asano, K.~Okada,
  K.~Kawasaki, and M.~Inaba, ``{Estimation and Control of Motor Core
  Temperature with Online Learning of Thermal Model Parameters: Application to
  Musculoskeletal Humanoids},'' \emph{IEEE Robotics and Automation Letters},
  vol.~5, no.~3, pp. 4273--4280, 2020.

\end{thebibliography}
}

\end{document}